\documentclass{article}

\usepackage{microtype}
\usepackage{graphicx}
\usepackage{subcaption}
\usepackage{longtable}
\usepackage{array} %
\usepackage{booktabs} %
\usepackage{hyperref}
\usepackage{xurl}
\usepackage{placeins}
\usepackage[table]{xcolor}
\usepackage{arydshln}
\usepackage[normalem]{ulem}   %
\usepackage{tabularx}
\usepackage{comment}
\definecolor{AvgShade}{HTML}{FBF6DE} %
\usepackage{pgfplots}
\pgfplotsset{compat=1.18}
\usepgfplotslibrary{fillbetween}
\usetikzlibrary{intersections}
\newcommand{\srciPct}[2]{#1{\footnotesize\,\textcolor{black!75}{$\pm$\,#2}}}
\usepackage{longtable}  %
\usepackage{array}      %
\usepackage{seqsplit}   %

\usepackage{booktabs}
\usepackage{array}
\usepackage{longtable}
\usepackage{xurl} %

\newcommand{\paramname}[1]{\path{#1}}

\usepackage[preprint]{icml2026}

\usepackage{amsmath}
\usepackage{amssymb}
\usepackage{mathtools}
\usepackage{amsthm}

\pgfplotsset{compat=1.18}

\usepackage[capitalize,noabbrev]{cleveref}

\theoremstyle{plain}

\theoremstyle{definition}

\theoremstyle{remark}

\usepackage[textsize=tiny]{todonotes}

\icmltitlerunning{Embedding Morphology into Transformers for Cross-Robot Policy Learning}

\begin{document}

\twocolumn[
  \icmltitle{Embedding Morphology into Transformers for Cross-Robot Policy Learning}

  \icmlsetsymbol{equal}{*}

  \begin{icmlauthorlist}
    \icmlauthor{Kei Suzuki}{comp}
    \icmlauthor{Jing Liu}{comp}
    \icmlauthor{Ye Wang}{comp}
    \icmlauthor{Chiori Hori}{comp}
    \icmlauthor{Matthew Brand}{comp}
    \icmlauthor{Diego Romeres}{comp}
    \icmlauthor{Toshiaki Koike-Akino}{comp}
  \end{icmlauthorlist}

  \icmlaffiliation{comp}{Mitsubishi Electric Research Laboratories (MERL), Cambridge, Massachusetts, USA}
  \icmlcorrespondingauthor{Kei Suzuki}{kesuzuki@merl.com}

  \icmlkeywords{vision-language-action, robot learning, multi-embodiment}

  \vskip 0.3in
]

\printAffiliationsAndNotice{}  %

\begin{abstract}
Cross-robot policy learning---training a single policy to perform well across multiple embodiments---remains a central challenge in robot learning. Transformer-based policies, such as vision-language-action (VLA) models, are typically embodiment-agnostic and must infer kinematic structure purely from observations, which can reduce robustness across embodiments and even limit performance within a single embodiment. We propose an embodiment-aware transformer policy that injects morphology via three mechanisms: (1) kinematic tokens that factorize actions across joints and compress time through per-joint temporal chunking; (2) a topology-aware attention bias that encodes kinematic topology as an inductive bias in self-attention, encouraging message passing along kinematic edges; and (3) joint-attribute conditioning that augments topology with per-joint descriptors to capture semantics beyond connectivity.
Across a range of embodiments, this structured integration consistently improves performance over a vanilla pi0.5 VLA baseline, indicating improved robustness both within an embodiment and across embodiments.
\end{abstract}

\begin{figure}[t]
  \centering
  \includegraphics[width=1.0\columnwidth]{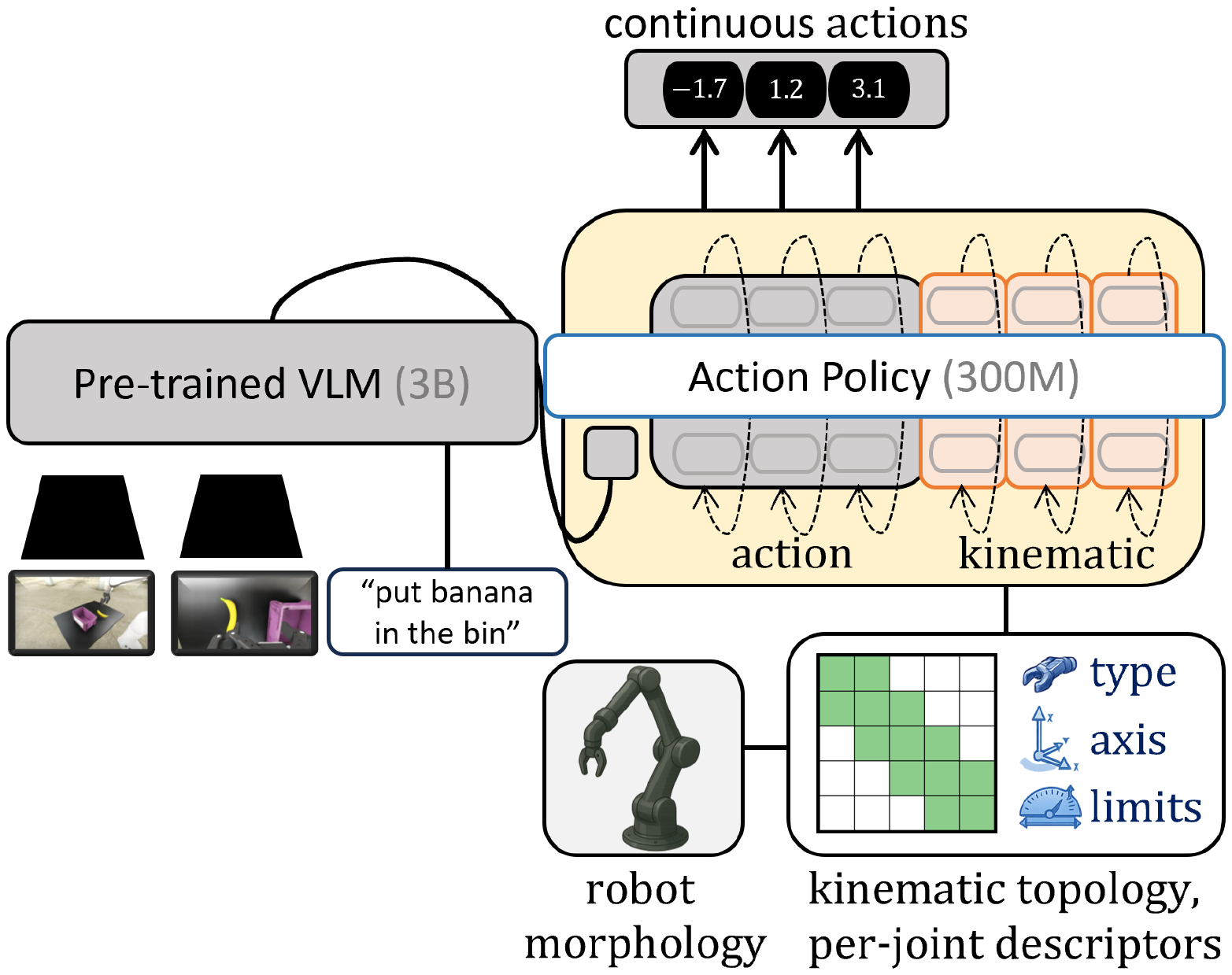}
  \caption{\textbf{Embedding robot morphology into a Transformer-based VLA policy}:
We embed kinematic topology and per-joint semantics into the action policy.
This design consistently improves cross-robot policy learning.}
  \label{fig:overview}
\end{figure}

\section{Introduction}
Transformer-based robot policies, especially vision-language-action (VLA) models \cite{ahn2022can,brohan2022rt,driess2023palm}, have advanced rapidly by scaling to large and diverse datasets, yielding increasingly generalist controllers that can execute a broad range of tasks.
However, \emph{cross-robot policy learning}---training a single policy to perform well across diverse robot embodiments---remains a central challenge.
This includes robustness to embodiment changes in the real world, such as hardware variation or failures, design upgrades, or deployment on entirely different robot platforms.
In practice, achieving strong performance across embodiments often requires additional data and training for each robot, e.g., via fine-tuning \cite{intelligence2504pi0,kim2024openvla} and/or replacing the action head \cite{team2024octo,doshi2024scaling} to match the target action space.
Such approaches reflect a structural limitation: many VLA policies are typically embodiment-agnostic and must learn kinematics and cross-joint coordination implicitly from observations, making cross-robot policy learning particularly challenging.
This motivates injecting embodiment as an explicit inductive bias into the policy architecture.
A common formulation represents robot morphology as a \emph{kinematic graph}, where nodes correspond to actuated joints and edges capture physical connectivity. Prior work incorporates this structure through graph neural networks \cite{wang2018nervenet,huang2020one,whitman2023learning} or topology-aware attention in transformers \cite{velivckovic2017graph,hong2021structure,sferrazza2024body}, improving cross-robot policy learning.
Nevertheless, existing embedding approaches face three challenges.
(i) \textbf{Lack of a kinematic token interface:}
VLA models such as $\pi_{0.5}$ \cite{intelligence2504pi0} compress joint-space structure into a compact set of action tokens, making it unclear where to apply existing morphology-embedding methods.
(ii) \textbf{Local--global trade-off in topology-aware attention:}
Enforcing strong locality promotes kinematic message passing but can limit long-range coordination.
(iii) \textbf{Missing joint semantics:} Existing methods do not capture per-joint semantics, even though joints with identical topology can play different functional roles (e.g., actuation type and limits).

\paragraph{Contributions}
To address these limitations, we propose an embodiment-aware transformer policy that injects robot morphology into the VLA action policy via three mechanisms:
(1) \emph{kinematic tokens} to provide a joint-wise action representation;
(2) \emph{topology-aware attention bias} with a local/global schedule to balance kinematic message passing and global context; and
(3) \emph{joint-attribute conditioning} to incorporate per-joint semantics beyond connectivity.
Across single- and multi-embodiment evaluations on DROID (Franka Panda), Unitree G1 Dex1, and SO101, our structured morphology encoding improves success rates over the vanilla $\pi_{0.5}$ VLA baseline \cite{intelligence2504pi0}, indicating improved robustness both within an embodiment and across embodiments.

\begin{figure*}[t]
  \centering
  \begin{subfigure}[b]{0.48\textwidth}
    \centering
    \includegraphics[width=1.00\linewidth]{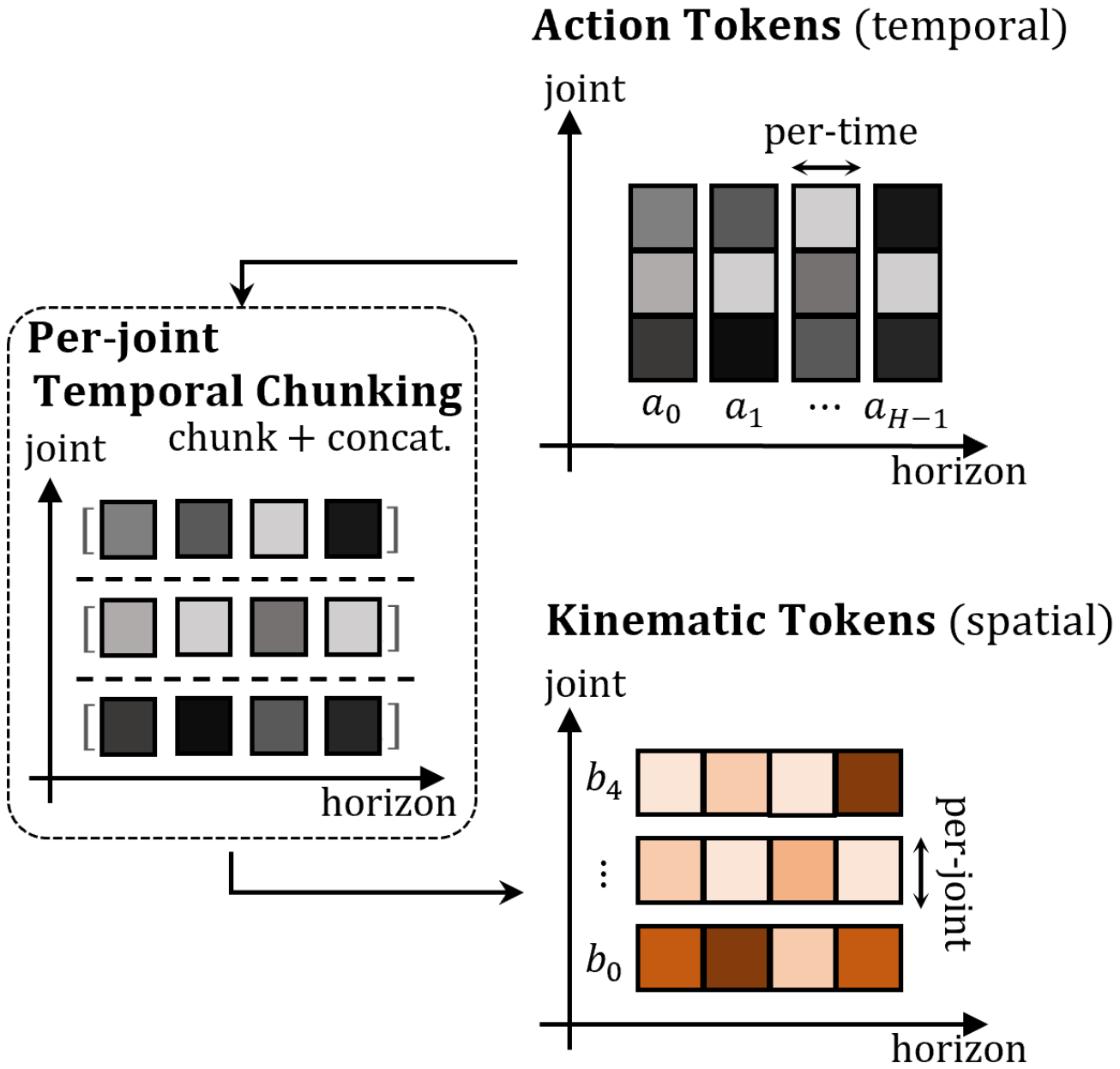}
    \subcaption{\textbf{Kinematic Token (KT)}:
Kinematic tokens provide a joint-wise interface for the VLA action policy.
While the standard action tokens retain temporal structure, kinematic tokens compress the horizon into per-joint summaries, emphasizing cross-joint (spatial) structure and enabling topology/semantics embedding.}
    \label{fig:temporal_chunk}
  \end{subfigure}
  \hfill
  \begin{subfigure}[b]{0.50\textwidth}
    \centering
    \includegraphics[width=\linewidth]{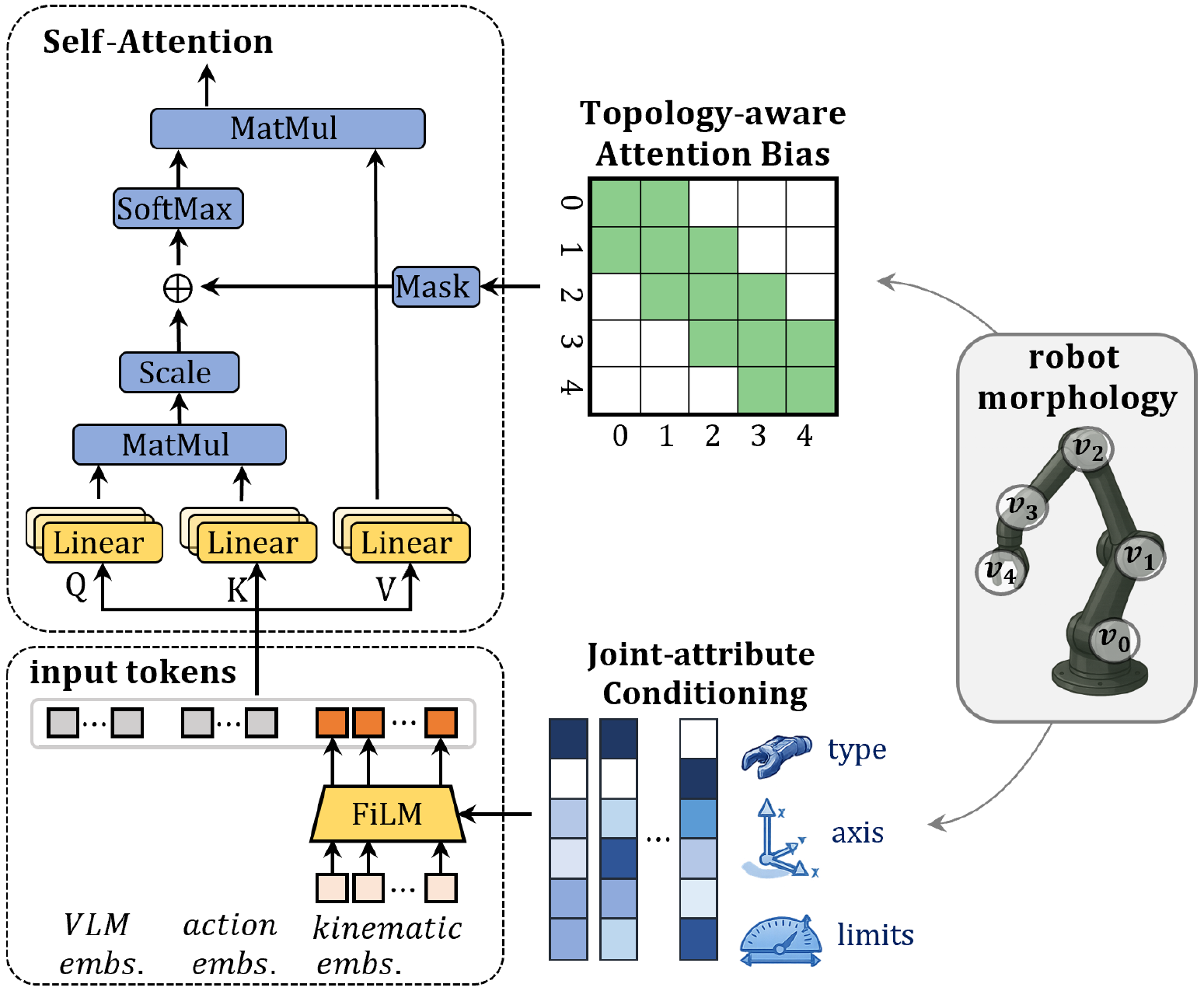}
    \subcaption{\textbf{Topology bias and joint-attribute conditioning}:
We embed kinematic topology and semantics in two ways:
(i) a topology bias encourages kinematic message passing by restricting joint-to-joint self-attention to connected joints, and
(ii) FiLM conditions kinematic-token embeddings on per-joint descriptors to disambiguate joint roles beyond connectivity.}
    \label{fig:morphology_injection}
  \end{subfigure}
  \caption{\textbf{Embodiment-aware Transformer Policy}:
Morphology is embedded via three mechanisms: (1) kinematic tokens; (2) topology-aware attention bias; and (3) joint-attribute conditioning.}
  \label{fig:method}
\end{figure*}

\section{Related Work}

\subsection{Data-driven generalist robot policies}
Recent progress in robot policy learning has been driven by scaling vision-language-action (VLA) models---large transformer policies that couple a pretrained VLM backbone with an action policy head \cite{ahn2022can,brohan2022rt,driess2023palm}.
A common recipe is to pretrain on large-scale robot datasets \cite{walke2023bridgedata,o2024open,khazatsky2024droid}, and then adapt to a target robot via fine-tuning or lightweight adaptation modules \cite{team2024octo,doshi2024scaling}.
Despite these advances, cross-robot policy learning---training a single policy to perform well across multiple
embodiments--- remains a central challenge, as 
these policies are typically embodiment-agnostic and must infer kinematic structure from observations alone.

\subsection{Architectures for embodiment-aware robot policies}
A common approach to improve cross-robot policy learning is to explicitly encode robot morphology as a \emph{kinematic graph}, where nodes correspond to actuated joints and edges represent physical connectivity. This explicit representation provides a structured interface for multi-embodiment learning.

\paragraph{GNN policies with kinematic-graph message passing}
Graph neural network (GNN) policies encode morphology by running message passing on a kinematic graph, promoting body-wide coordination \cite{wang2018nervenet,huang2020one,whitman2023learning}.
A key design question in this line of work is how information should propagate over the kinematic graph---balancing local interactions with global coordination.
NerveNet \cite{wang2018nervenet} applies repeated local (one-hop) message passing so that information can gradually propagate across the body.
SMP \cite{huang2020one} further explores structured propagation by separating bottom-up and top-down information flow along the kinematic tree, aiming to capture hierarchical coordination patterns.
Despite these advances, designing message-passing schemes that simultaneously support \emph{local} and \emph{global} information propagation remains challenging.

\paragraph{Transformer policies with topology-aware attention}
Transformer-based policies facilitate both local and global information exchange by self-attention, and inject kinematic priors by incorporating the kinematic graph directly into the attention mechanism \cite{velivckovic2017graph,hong2021structure,sferrazza2024body}.
Concretely, these methods use the kinematic graph to modulate attention, either by \emph{Hard-Mask} attention or \emph{Soft-Mask} attention.
Hard-masking approaches \cite{sferrazza2024body,buterez2024masked} inject topology by enforcing a binary attention constraint:
each kinematic token can attend only to itself and its 1-hop kinematic neighbors in the kinematic graph, and all other joint pairs are disallowed (implemented by masking their attention logits before the softmax).
To recover global coordination, some mixed designs alternate topology-masked (local) layers with fully connected (global) attention layers.
Soft-Mask attention approaches \cite{hong2021structure,luo2025gcnt} inject topology by keeping full attention but adding a learnable, topology-conditioned bias term to the attention logits, so kinematically closer joints receive higher attention on average without strictly blocking any pairs.
Although Soft-Mask approaches offer greater flexibility for embedding kinematics, prior works \cite{sferrazza2024body,buterez2024masked} report optimization instability, whereas Hard-Mask designs tend to be more stable and can outperform Soft-Mask variants.
In our method, we follow this taxonomy and consider both families of topology-aware attention within a unified framework (Full-/Mix-Mask for Hard-Mask; SPD-based bias for Soft-Mask).
Despite these advances, two limitations remain:
\begin{itemize}
    \item \textbf{Limited applicability to VLA token interfaces:} In state-of-the-art VLA models such as $\pi_{0.5}$ \cite{intelligence2504pi0}, action generation often compresses joint-space structure into a small set of action tokens, making it difficult to apply existing kinematic biasing schemes at the level of individual joints.
    \item \textbf{Topology alone lacks joint semantics:} Connectivity specifies how information propagates, but not what each joint represents. Encoding per-joint semantics is important for disambiguating functional roles and enabling more structured action generation (e.g., actuation type, axis, or joint limits).
\end{itemize}

\section{Proposed Method}
We propose an embodiment-aware transformer policy that injects robot morphology into the VLA action policy via three mechanisms: (1) kinematic tokens for factorized joint-action representation; (2) a topology-aware attention bias for kinematic message passing; and (3) joint-attribute conditioning for semantics beyond connectivity.
An overview is depicted in Figure~\ref{fig:method}.

\subsection{Kinematic Tokens (KT)}
To factorize the action sequence across joints while compressing temporal information, we introduce \emph{kinematic tokens} in addition to the standard action tokens used in VLA policies. The action tokens retain fine-grained temporal structure, while the kinematic tokens provide a compact, per-joint view that highlights cross-joint structure (Figure~\ref{fig:temporal_chunk}).

Let $a_{t,j}$ denote the scalar action for joint $j\in\{0,\ldots, J-1\}$ at time $t \in \{0,\ldots,H-1\}$, where $J$ and $H$ are the maximum number of joints and the maximum horizon length, respectively. 
The original $\pi_{0.5}$ VLA uses action tokens coupling all joints into one embedding per horizon, $[a_{t,j}]_{j=0}^{J-1}$, that prevents the use of topology-aware attention.
To embed morphology, we decouple the action into the spatial domain.
We then split the horizon into $G$ non-overlapping temporal chunks of size $g = \frac{H}{G}$. 
For each joint $j$, we concatenate the $g$ horizon actions within the chunk into a vector
\begin{equation}
b_{j,T_k} := [a_{t,j}]_{t\in T_k} \in \mathbb{R}^{g},
\end{equation}
where for each chunk $k \in \mathcal{K} := \{0,1,\ldots,G-1\}$, we define the index set
\begin{equation}
T_k = \{kg,\; kg+1,\; \ldots,\; (k+1)g-1\}.
\end{equation}
We refer to the vector $b_{j,T_k}$ as the \emph{kinematic token} for joint $j$ and chunk $k$.
This yields $G$ kinematic tokens per joint (and $JG$ kinematic tokens in total for $J$ joints). 
We project each kinematic token $b_{j,T_k}$ to a $d$-dimensional embedding
$z_{j,T_k}=\mathrm{Enc}_0(b_{j,T_k}) \in \mathbb{R}^d$ with a lightweight multi-layer perceptron (MLP), and append them to the token sequence as additional context for the action expert of VLA.
Following the VLA backbone, the policy predicts actions from the action tokens, which can attend to these \emph{kinematic tokens} to leverage joint-wise structure.
In our experiments, we find that using a single chunk ($G{=}1$) performs best.

\paragraph{Auxiliary kinematic tokens (AKT)}
Building on the kinematic token interface, we further increase the token capacity per joint by introducing
\emph{auxiliary kinematic tokens}.
Specifically, for each kinematic token $b_{j,T_k}$, in addition to the standard embedding
$z_{j,T_k}=\mathrm{Enc}_0(b_{j,T_k}) \in \mathbb{R}^d$, we generate $M$ auxiliary embeddings
\begin{equation}
z^{(m)}_{j,T_k}=\mathrm{Enc}_m(b_{j,T_k}) \in \mathbb{R}^d,\quad m=1,\ldots,M,
\end{equation}
using lightweight encoders with the same input but independent parameters.
We then append these auxiliary tokens to the kinematic token sequence, so the transformer can attend to a richer
set of per-joint representations.

\subsection{Topology-aware attention}
Vanilla self-attention treats the token sequence as a fully connected graph, allowing arbitrary interactions between joints.
In contrast, robot embodiments admit a natural kinematic topology represented as a graph.
Let $\mathcal{G}=(\mathcal{V},\mathcal{E})$ be the kinematic graph, where each vertex $v_j\in\mathcal{V}$ corresponds to joint $j$, and each edge $(v_i,v_j)\in\mathcal{E}$ indicates physical connectivity.
We encode this topology as an inductive bias in self-attention (Figure~\ref{fig:morphology_injection}).

\paragraph{Joint-to-joint self-attention modulation}
In the VLA token sequence, we inject kinematic structure only within the \emph{joint-to-joint} attention block, while all other attention patterns remain identical to the original $\pi_{0.5}$ mask (Appendix~\ref{fig:attention_mask}).
Let $\mathrm{logits}^{(\ell)}_{i,j}$ denote the scaled dot-product attention logits at layer $\ell$ between query joint $i$ and key joint $j$.
We modulate joint-to-joint attention by adding a topology-dependent term $B^{(\ell)}_{i,j}$:
\begin{equation}
\alpha^{(\ell)}_{i,j}
=
\mathrm{softmax}_j\!\left(
\mathrm{logits}^{(\ell)}_{i,j} + B^{(\ell)}_{i,j}
\right),
\label{eq:topo_modulation_general}
\end{equation}
where $\alpha^{(\ell)}_{i,j}$ is the attention weight.

We consider three topology-bias variants in two families: Hard-Mask (Full-Mask and Mix-Mask) and Soft-Mask.
Hard-Mask uses an adjacency-based \emph{hard} bias that blocks attention to non-neighbor joints.
Soft-Mask uses a \emph{shortest-path distance} (SPD)-based \emph{soft} bias that favors nearby joints while retaining full attention.

\paragraph{Hard masking family:}
We first define the 1-hop neighborhood indicator
\begin{equation}
M_{i,j} =
\begin{cases}
1, & (i=j)\ \text{or}\ (i,j)\in\mathcal{E},\\
0, & \text{otherwise}.
\end{cases}
\label{eq:adj_indicator}
\end{equation}
Hard masking is implemented by setting
\begin{equation}
B^{(\ell)}_{i,j} =
\begin{cases}
0, & M_{i,j}=1,\\
-\infty, & M_{i,j}=0,
\end{cases}
\label{eq:topo_mask}
\end{equation}
so that non-neighbor joints receive zero attention after the softmax.

\paragraph{Full-Mask:}
Full-Mask applies hard masking Eq.~\eqref{eq:topo_mask} at every layer, enforcing strictly local (1-hop) joint-to-joint interactions.

\paragraph{Mix-Mask:}
Mix-Mask alternates masked and unmasked layers to balance local message passing with periodic global coordination.
Even-numbered layers apply hard masking (Eq.~\eqref{eq:topo_mask}), while odd-numbered layers use full attention ($B^{(\ell)}_{i,j}=0$ for all $i,j$).

\paragraph{Soft-Mask:}
We define the shortest-path distance on $\mathcal{G}$ as
\begin{equation}
d(i,j) = \min_{p:i\to j} |p|,
\label{eq:spd_def}
\end{equation}
where the minimum is taken over all paths $p$ from joint $i$ to joint $j$ and $|p|$ is the number of edges in the path (with $d(i,i)=0$).
Following a Graphormer-style design~\cite{ying2021transformers}, we parameterize the topology term by a learnable bias table indexed by distance:
\begin{equation}
B^{(\ell)}_{i,j} = \theta^{(\ell)}_{d(i,j)}.
\label{eq:topo_bias_spd}
\end{equation}
Unlike hard masking, Soft-Mask biases attention by kinematic distance while retaining full attention paths.

\begin{table}[t]
\centering
\caption{\textbf{Per-joint descriptors for joint-attribute conditioning}:
Each joint $j$ is represented by a descriptor $s_j$, which is used for FiLM-based conditioning of kinematic-token embeddings.
The descriptor includes joint type indicators, axis direction, motion limits, and contact-related properties (log-transformed where indicated); the \textit{Example} column shows a representative instance from the DROID (Franka Panda arm) embodiment.}
\small
\setlength{\tabcolsep}{5pt}
\renewcommand{\arraystretch}{1.15}
\begin{tabularx}{\columnwidth}{@{} l X r @{}}
\toprule
\textbf{Feature} & \textbf{Description} & \textbf{Example} \\
\midrule
\texttt{type\_pris}      & 1 if prismatic joint, else 0                 & 0        \\
\texttt{type\_rev}       & 1 if revolute joint, else 0                  & 1        \\
\texttt{ax}              & Joint axis X component (unit vector)         & 0        \\
\texttt{ay}              & Joint axis Y component (unit vector)         & 0        \\
\texttt{az}              & Joint axis Z component (unit vector)         & 1        \\
\texttt{hard\_lower}     & Hard lower limit [rad or m]                  & -2.9671  \\
\texttt{hard\_upper}     & Hard upper limit [rad or m]                  & 2.9671   \\
\texttt{damping\_log}    & $\log(\text{contact damping})$               & 6.90776  \\
\texttt{friction\_anchor}& Friction anchor flag (0/1)                   & 1        \\
\texttt{lateral\_friction}& Lateral friction coefficient                & 1        \\
\texttt{spinning\_friction}& Spinning friction coefficient              & 0.1      \\
\texttt{stiffness\_log}  & $\log(\text{contact stiffness})$             & 10.30895 \\

\bottomrule
\end{tabularx}
\label{tab:joint_features}
\end{table}

\begin{figure*}[t]
  \centering
  \scalebox{0.750}{%
    \begin{subfigure}[t]{0.32\textwidth}
      \centering
      \includegraphics[width=0.982\linewidth]{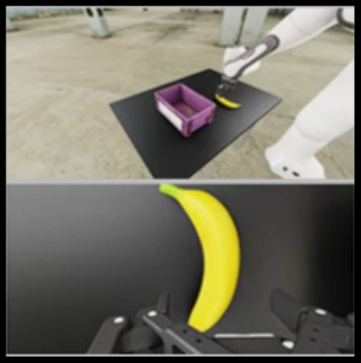}
      \caption{\textbf{DROID (Franka Panda):} Three pick-and-place tasks:
      (1) ``put the cube in the bowl'';
      (2) ``put the can in the mug'';
      (3) ``put banana in the bin''}
      \label{fig:sim_droid}
    \end{subfigure}
    \hspace{0.10\textwidth}
    \begin{subfigure}[t]{0.32\textwidth}
      \centering
      \includegraphics[width=\linewidth]{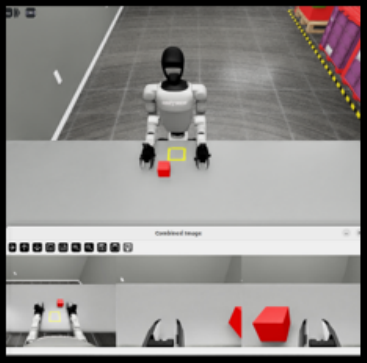}
      \caption{\textbf{Unitree G1 Dex1 (Unitree G1):} ``Pick up the red block and place it inside the yellow box.''}
      \label{fig:sim_unitree}
    \end{subfigure}
    \hspace{0.10\textwidth}
    \begin{subfigure}[t]{0.32\textwidth}
      \centering
      \includegraphics[width=\linewidth]{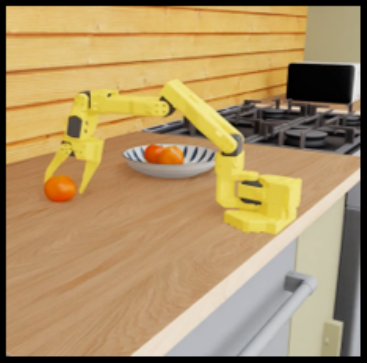}
      \caption{\textbf{SO101:} ``Grab orange and place into plate''}
       \label{fig:sim_so101}
    \end{subfigure}
  }
  \caption{\textbf{Simulation environments for evaluation}:
All environments are evaluated language-conditioned pick-and-place manipulation.}
  \label{fig:three_pdfs}
\end{figure*}

\subsection{Joint-attribute conditioning}
Topology-aware attention specifies \emph{which} joints can exchange information based on kinematic connectivity, but connectivity alone does not capture the \emph{semantics} of each joint (e.g., joints with similar local topology may play different functional roles). To complement topology, we condition kinematic-token embeddings on per-joint descriptors derived from robot morphology (Figure~\ref{fig:morphology_injection}).

For each joint $j$, we define a per-joint descriptor $s_j$ using the features in Table~\ref{tab:joint_features}, capturing the joint type, axis direction, motion limits, and contact-related properties. We use Feature-wise Linear Modulation (FiLM) \cite{perez2018film} to map $s_j$ to feature-wise scale and shift parameters:
\begin{equation}
\gamma_j,\beta_j = \mathrm{FiLM}(s_j),
\end{equation}
where $\gamma_j,\beta_j \in \mathbb{R}^{d}$. Given the kinematic-token embedding before conditioning $z_j \in \mathbb{R}^{d}$, we apply feature-wise affine modulation:
\begin{equation}
\tilde{z}_j = (1+\gamma_j)\odot z_j + \beta_j,
\end{equation}
where $\odot$ denotes element-wise multiplication. The conditioned embedding $\tilde{z}_j$ augments topology-based message passing with joint-specific semantics beyond connectivity.

\section{Experiments}
We evaluate morphology-aware transformer policies for \emph{cross-robot policy learning}, where a single policy is trained and evaluated on the same set of robot embodiments.
Our experiments isolate the contribution of each component:
(i) whether kinematic tokens improve success rates over the vanilla $\pi_{0.5}$ baseline;
(ii) whether topology-aware bias improves performance over unmasked attention;
(iii) whether joint-attribute conditioning provides additional gains over topology-only variants;
and (iv) whether these effects persist under single- and multi-embodiment training.

\subsection{Single-embodiment evaluations}
We first study single-embodiment imitation learning on two benchmarks to test whether explicit morphology encoding improves performance even outside cross-embodiment settings.
\textbf{DROID (Franka Panda):}
We use the public DROID dataset~\cite{khazatsky2024droid}, which contains demonstrations collected on an 8-DoF (including the gripper) Franka Panda arm.
We train on a 1/8 subset of DROID and evaluate in a simulation evaluation suite consisting of three language-conditioned pick-and-place tasks\footnote{\url{https://github.com/arhanjain/sim-evals}} (Figure~\ref{fig:sim_droid}).
\textbf{Unitree G1 Dex1 (Unitree G1):}
We also evaluate on the Unitree Dex1 benchmark in simulation, which focuses on manipulation with the Unitree G1 platform while the lower body remains stationary.
We train on the public Unitree Dex1 dataset with 16-DoF joint-space actions\footnote{\url{https://huggingface.co/datasets/unitreerobotics/G1_Dex1_PickPlaceRedBlock_Dataset_Sim}} and evaluate in an IsaacLab-based simulator\footnote{\url{https://github.com/unitreerobotics/unitree_sim_isaaclab}} (Figure~\ref{fig:sim_unitree}).

\begin{table}[t]
\centering
\caption{\textbf{Training protocol}:
Fine-tuning configurations and key hyperparameters for each benchmark.
All runs start from \texttt{pi05-base} with a cosine learning-rate schedule.
\textbf{AP-FT} updates only the action-policy components, whereas \textbf{Full-FT} updates the entire model.}
\small
\setlength{\tabcolsep}{4pt}
\renewcommand{\arraystretch}{1.1}
\begin{tabular}{l l r r r}
\toprule
\textbf{Setting} & \textbf{Fine-tuning} & \textbf{Batch} & \textbf{Horizon} & \textbf{Steps} \\
\midrule
DROID (Panda)    & AP-FT ($\sim$450M)   & 32 & 16 & 100k \\
Unitree G1 Dex1  & Full-FT ($\sim$3.5B)   & 8  & 32 & 60k \\
DROID+SO101      & AP-FT ($\sim$450M)   & 32 & 16 & 125k \\
\bottomrule
\end{tabular}
\label{tab:training_protocol}
\end{table}

\begin{table*}[t]
\centering
\caption{\textbf{Single-embodiment results on DROID}:
We evaluate whether embedding robot morphology improves performance even under single-embodiment training on DROID (Franka Panda).
Starting from the $\pi_{0.5}$ baseline, we add kinematic tokens, topology-aware attention, and joint-attribute conditioning (FiLM), individually and in combination.
Each component contributes to higher success rates, and the best performance is achieved by combining all three (KT + Mix-Mask + FiLM).
\textbf{Bold}: best SR in each column (ties allowed). \uline{Underline}: second-best. 95\% CI is reported with $\pm \Delta$.}
{%
\begin{tabular}{c l c r r r r}
\toprule
Kinematic Token & Mask & FiLM &
\multicolumn{4}{c}{Success Rate (SR\%) $\uparrow$ (95\% CI)} \\
\cmidrule(lr){4-7}
(Chunk $G$=1) &  &  &
Avg & Task 1 & Task 2 & Task 3 \\
\midrule

--  & --   & -- &
\cellcolor{AvgShade}\srciPct{19.7}{4.5} &
\srciPct{18.3}{4.4} & \srciPct{15.7}{4.1} & \srciPct{25.0}{4.9}\\

\cdashline{1-7}[0.6pt/2pt]

\checkmark  & --   & -- &
\cellcolor{AvgShade}\srciPct{36.0}{5.4} &
\srciPct{10.3}{3.6} & \srciPct{\uline{67.7}}{5.3} & \srciPct{30.0}{5.2}\\

\checkmark  & Full-Mask & -- &
\cellcolor{AvgShade}\srciPct{30.1}{5.2} &
\srciPct{15.3}{4.1} & \srciPct{34.0}{5.3} & \srciPct{41.0}{5.5}\\

\checkmark  & Mix-Mask  & -- &
\cellcolor{AvgShade}\srciPct{36.9}{5.4} &
\srciPct{\uline{27.7}}{5.0} & \srciPct{56.7}{5.6} & \srciPct{26.3}{5.1}\\

\checkmark  & Soft-Mask & -- &
\cellcolor{AvgShade}\srciPct{26.1}{4.9} &
\srciPct{\textbf{29.7}}{5.1} & \srciPct{17.3}{4.3} & \srciPct{31.3}{5.2}\\

\checkmark & --  & \checkmark &
\cellcolor{AvgShade}\srciPct{\uline{37.7}}{5.5} &
\srciPct{6.0}{2.7} & \srciPct{65.0}{5.4} & \srciPct{\uline{42.0}}{5.6} \\

\checkmark & Mix-Mask & \checkmark &
\cellcolor{AvgShade}\srciPct{\textbf{47.4}}{5.6} &
\srciPct{5.7}{2.7} & \srciPct{\textbf{77.7}}{4.7} & \srciPct{\textbf{58.7}}{5.5} \\

\bottomrule
\end{tabular}%
}
\label{tab:results1}
\end{table*}

\subsection{Multi-embodiment evaluation}
We evaluate multi-embodiment imitation learning by jointly training a single policy on a mixture of Panda and SO101 demonstrations, where the two robots differ in joint-space action dimensionality (Panda: 8-DoF; SO101: 6-DoF). This setting is challenging because the policy must reconcile embodiment-specific kinematics and coordination patterns under mismatched action spaces, and naive joint training can lead to interference between embodiments.
For the Panda portion, we use the same 1/8 subset of DROID and evaluation suite as in the single-embodiment setting. For SO101, we use the LeRobot dataset\footnote{\url{https://huggingface.co/datasets/LightwheelAI/leisaac-pick-orange}} and evaluate in the corresponding LeIsaac simulator\footnote{\url{https://github.com/LightwheelAI/leisaac}} (Figure~\ref{fig:sim_so101}). 
During training, each mini-batch contains trajectories from a single embodiment, and we sample Panda and SO101 batches with an 8:2 ratio (i.e., 80\% Panda and 20\% SO101).
We choose this ratio to keep the same amount of DROID (Panda) data as in our single-embodiment setting, while SO101 can be trained adequately with a smaller dataset in this regime.

\subsection{Metrics}
We evaluate policies using task \emph{success rate} (SR), computed over 300 rollout trials per experimental condition.
Success is determined by an axis-aligned bounding box (AABB) criterion: an episode is successful if the object remains inside the target region for a minimum dwell time and is approximately stationary.
To quantify statistical uncertainty, we also report a 95\% confidence interval (CI) for each SR using the Wilson score interval \cite{brown2001interval} for a binomial proportion ($n{=}300$).
In tables, we present results in the format $\mathrm{SR}\ \pm\ \Delta$, where $\Delta$ denotes the CI half-width, i.e., $\Delta = (\mathrm{CI}_{\mathrm{upper}}-\mathrm{CI}_{\mathrm{lower}})/2$.
Across trials, we vary the random seed to change initial conditions (e.g., object placement).

\subsection{Model variants}
We evaluate variants built on the $\pi_{0.5}$ backbone by adding kinematic tokens, topology-aware attention, and joint-attribute conditioning, both individually and in combination.
For topology-aware attention, we compare Full-Mask, Mix-Mask and Soft-Mask variants.
Kinematic tokens are embedded with a linear--SwiGLU--linear MLP, and joint-attribute conditioning uses a linear FiLM generator. In addition, we ablate the temporal chunk size ($G\in\{1,2,4,8,16\}$) and the kinematic token capacity by introducing auxiliary kinematic tokens.

\subsection{Training protocol}
All variants are initialized from the $\pi_{0.5}$ checkpoint \texttt{pi05-base}\footnote{\url{gs://openpi-assets/checkpoints/pi05_base}}. For architectures that add morphology modules, we zero-initialize the final layer of each module so that it starts near an identity mapping, stabilizing optimization.
For each benchmark, our \textbf{baseline} is vanilla $\pi_{0.5}$, fine-tuned from \texttt{pi05-base} on that benchmark.
Depending on computational budget, we use either action policy fine-tuning (\textbf{AP-FT}) which updates only the diffusion action expert, or full fine-tuning (\textbf{Full-FT}) which updates the entire model.
Figure~\ref{tab:training_protocol} summarizes the training configuration for each setting.

\section{Results}

\subsection{Single-embodiment results on DROID}
Table~\ref{tab:results1} shows \emph{average success rate} (Avg SR) for $\pi_{0.5}$ variants, including vanilla baseline.
Overall, structured morphology encoding improves performance, and the best result is achieved by combining kinematic Tokens with Mix-Mask and joint-attribute conditioning.
Below, we isolate the contribution of (1) kinematic tokens, (2) topology-aware attention bias, and (3) joint-attribute conditioning. For completeness, Appendix~\ref{detailed_DROID} reports \emph{success time} on DROID as well as an ablation over encoder variants.

\textbf{(1) kinematic tokens}:
Compared to the baseline (Avg SR 19.7\%), adding kinematic tokens improves Avg SR to 36.0\%, showing that a joint-centric action representation is effective even without explicit topology or semantics.

\textbf{(2) Topology-aware attention}:
Starting from the kinematic token model (Avg SR 36.0\%), Mix-Mask improves Avg SR to 36.9\%, suggesting that alternating topology-constrained (local) and full (global) attention is beneficial.
In contrast, Full-Mask reduces Avg SR to 30.1\%, indicating that enforcing 1-hop locality at every layer can be overly restrictive.
In addition, Soft-Mask achieves Avg SR 26.1\%, which is lower than both masked variants.

\textbf{(3) Joint-attribute conditioning}:
Adding FiLM-based joint-attribute conditioning yields consistent gains.
Without topology encoding, FiLM improves Avg SR from 36.0\% to 37.7\%, and with Mix-Mask Avg SR is improved from 36.9\% to 47.4\%, achieving the best overall performance.
Notably, the full morphology encoding (KT+Mix+FiLM) yields significantly large gains over the baseline on Task~2 and Task~3, improving success rates by 5-fold and 2.3-fold, respectively.

\begin{table}[t]
\centering
\caption{\textbf{Single-embodiment results on Unitree G1 Dex1}:
We evaluate our method on Unitree G1 Dex1 to test its benefit beyond DROID.
A key difference is the 16-DoF joint-space action space.
Overall, our components remain effective in this setting, and combining KT with Mix-Mask and FiLM achieves the best SR.
\textbf{Bold}: best SR. \uline{Underline}: second-best.}
\setlength{\tabcolsep}{4pt} %
\begin{tabular}{c l l r}
\toprule
Kinematic Token & Mask & FiLM & Avg SR\%$\uparrow$ \\
(Chunk $G$=1) &  &  & (95\% CI) \\
\midrule
-- & --   & --         & \cellcolor{AvgShade}\srciPct{24.7}{4.9} \\

\cdashline{1-4}[0.6pt/2pt]

\checkmark & --   & --         & \cellcolor{AvgShade}\srciPct{24.3}{4.9} \\
\checkmark & Full-Mask & --         & \cellcolor{AvgShade}\srciPct{23.8}{4.8} \\
\checkmark & Mix-Mask  & --         & \cellcolor{AvgShade}\srciPct{\uline{27.3}}{5.0} \\

\checkmark & --  & \checkmark & \cellcolor{AvgShade}\srciPct{25.3}{4.9} \\
\checkmark & Mix-Mask & \checkmark & \cellcolor{AvgShade}\srciPct{\textbf{28.0}}{5.0} \\

\bottomrule
\end{tabular}%
\label{tab:film_scene1_sr}
\end{table}

\begin{figure}[t]
  \centering
  \begin{tikzpicture}
    \begin{axis}[
      width=\columnwidth,
      height=0.62\columnwidth,
      xlabel={Training steps},
      scaled x ticks=false,
      ylabel={Macro SR},
      xmin=0, xmax=125000,
      ymin=0, ymax=0.35,
      xtick={0,50000,80000,125000},
      xticklabels={0,50k,80k,125k},
      ytick={0,0.1,0.2,0.3},
      grid=both,
      major grid style={line width=0.2pt, draw=black!12},
      minor grid style={line width=0.1pt, draw=black!6},
      tick align=outside,
      tick pos=left,
      legend style={at={(0.02,0.98)},anchor=north west,draw=none,fill=none},
      legend cell align=left,
      every axis plot/.append style={line width=1.9pt, mark size=3.4pt},
    ]

      \addplot[name path=pi05_upper, draw=none, forget plot]
        coordinates {(0,0.0000) (50000,0.0454) (80000,0.0700) (125000,0.1747)};
      \addplot[name path=pi05_lower, draw=none, forget plot]
        coordinates {(0,0.0000) (50000,0.0136) (80000,0.0287) (125000,0.1072)};
      \addplot[gray!60, fill opacity=0.18, draw=none, forget plot]
        fill between[of=pi05_upper and pi05_lower];

      \addplot+[color=gray!75, dashed, mark=o]
        coordinates {(0,0.0000) (50000,0.0250) (80000,0.0450) (125000,0.1375)};
      \addlegendentry{$\pi_{0.5}$}

      \addplot[name path=ours_upper, draw=none, forget plot]
        coordinates {(0,0.0000) (50000,0.2233) (80000,0.2206) (125000,0.2526)};
      \addplot[name path=ours_lower, draw=none, forget plot]
        coordinates {(0,0.0000) (50000,0.1477) (80000,0.1454) (125000,0.1729)};
      \addplot[orange!70, fill opacity=0.18, draw=none, forget plot]
        fill between[of=ours_upper and ours_lower];

      \addplot+[color=orange, solid, mark=diamond*]
        coordinates {(0,0.0000) (50000,0.1825) (80000,0.1800) (125000,0.2100)};
      \addlegendentry{Ours (KT+Mix+FiLM)}

    \end{axis}
  \end{tikzpicture}
  \caption{\textbf{Multi-embodiment learning curves on Panda--SO101}:
We jointly train a single policy on a mixed Panda (DROID) and SO101 dataset.
We report Macro SR, defined as $(\mathrm{SR}_{\text{Panda}}+\mathrm{SR}_{\text{SO101}})/2$. Our full model outperforms the $\pi_{0.5}$ baseline throughout training. For completeness, per-embodiment success rates are reported in the appendix~\ref{app:multi}. Shaded regions indicate 95\% confidence intervals.}
  \label{fig:sr_vs_steps_macro}
\end{figure}
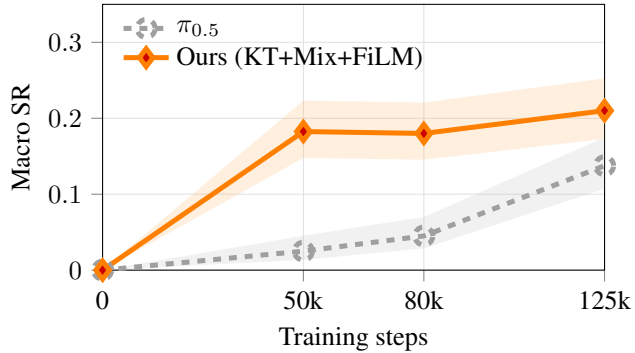

\subsection{Single-embodiment results on Unitree G1 Dex1}
Table~\ref{tab:film_scene1_sr} shows Avg SR for each variant on Unitree G1 Dex1.
Consistent with DROID, the full morphology encoding (KT, Mix-Mask and FiLM) achieves the best performance (Avg SR 28.0\%).
Topology-aware attention masking and joint-attribute conditioning provide additional gains on top of kinematic tokens: Mix-Mask improves Avg SR from 24.3\% to 27.3\%, and adding FiLM further improves to 28.0\%.

\subsection{Multi-embodiment results on DROID + SO101}
Figure~\ref{fig:sr_vs_steps_macro} shows learning curves for multi-embodiment training on the Panda/SO101 mixture. We report Macro SR, defined as $(\mathrm{SR}_{\text{DROID}}+\mathrm{SR}_{\text{SO101}})/2$.
Overall, our embodiment-aware policy (KT, Mix-Mask, and FiLM) achieves higher Avg SR throughout training. For example, at 50k steps our method reaches Avg SR of 15.5\% while $\pi_{0.5}$ has a poor SR of 5.0\%, and at 125k steps our method remains higher (20.7\% vs.\ 17.5\%).

\subsection{Ablation study}
This section examines key design choices in our embodiment-aware Transformer and further probes a strong alternative for topology-aware attention.
We study whether performance depends on (i) the kinematic token chunk size $G$, (ii) auxiliary kinematic tokens (AKT), and (iii) bias initialization for the Soft-Mask variant, which is potentially more expressive.

\begin{table}[t]
\centering
\caption{\textbf{Effect of temporal chunk size on DROID}:
We study how the temporal granularity of kinematic tokens affects performance.
Using a single chunk ($G{=}1$) achieves the best performance, while larger chunk sizes reduce performance.
\textbf{Bold}: best SR. \uline{Underline}: second-best.}
\begin{minipage}{0.70\columnwidth}
\centering
\setlength{\tabcolsep}{4pt}
\renewcommand{\arraystretch}{1.05}
\begin{tabular}{c r}
\toprule
Kinematic Tokens & \shortstack[c]{Avg SR\%$\uparrow$} \\
Chunk ($G$) & (95\% CI)\\
\midrule
1  & \cellcolor{AvgShade}\srciPct{\textbf{36.0}}{5.4} \\
2  & \cellcolor{AvgShade}\srciPct{\uline{35.8}}{5.4} \\
4  & \cellcolor{AvgShade}\srciPct{34.4}{5.3} \\
8  & \cellcolor{AvgShade}\srciPct{30.5}{5.2} \\
16 & \cellcolor{AvgShade}\srciPct{33.3}{5.3} \\
\bottomrule
\end{tabular}
\end{minipage}
\label{tab:pi05_selected_temporal_chunks}
\end{table}

\paragraph{Effect of temporal chunk size:}
To study how the temporal resolution of kinematic tokens affects performance, we evaluated five temporal chunk sizes ($G\in\{1,2,4,8,16\}$), where smaller $G$ corresponds to more aggressive temporal compression.
Table~\ref{tab:pi05_selected_temporal_chunks} shows that using a single chunk ($G{=}1$) achieves the best Avg SR  of 36.0\%, while performance generally degrades as $G$ increases (33.3\% at $G{=}16$).

\paragraph{Effect of auxiliary kinematic tokens (AKT):}
We test whether increasing per-kinematic token capacity improves performance via auxiliary kinematic tokens (AKT).
Table~\ref{tab:results_akt_ablation} shows that AKT consistently improves Avg SR across configurations.
Notably, under Mix-Mask, AKT increases Avg SR from 37.0\% to 47.3\%, indicating that scaling kinematic token capacity can substantially boost performance.

\begin{table}[t]
\centering
\caption{\textbf{Effect of auxiliary kinematic tokens (AKT) on DROID}:
We study whether increasing per-joint token capacity improves performance.
Adding AKT consistently improves performance, especially with Mix-Mask.
\textbf{Bold}: best SR. \uline{Underline}: second-best.}
\begin{minipage}{0.70\columnwidth}
\centering
\setlength{\tabcolsep}{4pt}
\renewcommand{\arraystretch}{1.05}
\begin{tabular}{l c r}
\toprule
\multicolumn{2}{c}{Kinematic Tokens} & \shortstack[c]{Avg SR\%$\uparrow$} \\
Mask & AKT & (95\% CI) \\
\midrule
--       & --         & \cellcolor{AvgShade}\srciPct{36.0}{5.4} \\
--       & \checkmark & \cellcolor{AvgShade}\srciPct{\uline{37.0}}{5.4} \\
Full-Mask& --         & \cellcolor{AvgShade}\srciPct{30.3}{5.2} \\
Full-Mask& \checkmark & \cellcolor{AvgShade}\srciPct{33.0}{5.3} \\
Mix-Mask & --         & \cellcolor{AvgShade}\srciPct{\uline{37.0}}{5.4} \\
Mix-Mask & \checkmark & \cellcolor{AvgShade}\srciPct{\textbf{47.3}}{5.6} \\
\bottomrule
\end{tabular}
\end{minipage}
\label{tab:results_akt_ablation}
\end{table}

\begin{table}[t]
\centering
\caption{\textbf{Effect of bias initialization on DROID}:
We study the effect of bias initialization in Soft-Mask attention (Zero, Hard, Mix, Linear).
Overall, Mix achieves the best performance among Soft-Mask variants, yet all Soft-Mask variants remain below the masked variants in Table~\ref{tab:results1}.
\textbf{Bold}: best. \uline{Underline}: second-best.}
\begin{minipage}{0.70\columnwidth}
\centering
\setlength{\tabcolsep}{6pt}
\begin{tabular}{l@{\hspace{10pt}}r}
\toprule
Bias Init & \shortstack[c]{Avg SR\%$\uparrow$\\(95\% CI)} \\
\midrule
Zero   & \cellcolor{AvgShade}\srciPct{\uline{26.1}}{4.9} \\
Hard   & \cellcolor{AvgShade}\srciPct{25.1}{4.9} \\
Mix    & \cellcolor{AvgShade}\srciPct{\textbf{28.1}}{5.1} \\
Linear & \cellcolor{AvgShade}\srciPct{20.4}{4.5} \\
\bottomrule
\end{tabular}
\end{minipage}
\label{tab:bias_init_ablation}
\end{table}

\paragraph{Effect of bias initialization:}
Since Soft-Mask attention is potentially more expressive yet optimization-sensitive, we test whether bias initialization improves its performance on DROID. All models use kinematic tokens ($G{=}1$) and no FiLM; only the initialization of the learnable bias table is changed.
We study four initializations that impose different topology priors at the start of training:
\begin{itemize}
    \item \textbf{Zero:} All bias parameters are initialized to zero, corresponding to no topology prior.
    \item \textbf{Hard:} biases are initialized to approximate a full-mask prior by assigning a negative bias (e.g., $-3$) to non-neighbor joint pairs.
    \item \textbf{Mix:} To mimic the alternating local/global schedule, even layers use the \textbf{Hard} initialization while odd layers are initialized to \textbf{Zero}.
    \item \textbf{Linear:} Biases are initialized as a distance-dependent prior, linearly interpolating from $0$ for near pairs to $-3$ for far pairs.
\end{itemize}

Table~\ref{tab:bias_init_ablation} shows the effect of bias initialization.
We found that Mix initialization performs best (Avg SR $28.1\%$).
Nevertheless, across all initializations, the Soft-Mask variant does not surpass the Hard-Masked variants (e.g., Mix-Mask in Table~\ref{tab:results1}). 
This is consistent to some related reports \cite{sferrazza2024body,buterez2024masked}.
In Appendix~\ref{app:softmask}, we further investigate Soft-Mask with more detailed variants, but we do not observe consistent improvements.

\section*{Conclusion}
We presented an embodiment-aware transformer policy that injects robot morphology through (1) kinematic tokens with per-joint temporal chunking, (2) a topology-aware attention mask, and (3) joint-attribute conditioning.
Across single- and multi-embodiment evaluations on DROID (Franka Panda), Unitree G1 Dex1, and SO101, our structured morphology encoding improves success rates over the vanilla $\pi_{0.5}$ VLA baseline, indicating improved robustness both within an embodiment and across embodiments.
In the future, we plan to (i) further study kinematic-token representations, focusing on token design choices and scalability with token capacity, (ii) develop reliable Soft-Mask attention methods with more stable optimization, and (iii) develop more efficient multi-embodiment training strategies---including curriculum schedules that better preserve single-embodiment performance while scaling to additional robots.

\FloatBarrier

\section*{Impact Statement}

Our work aims to enable robot policies that perform well across diverse robot embodiments by explicitly encoding robot morphology.
This fits within a broader research area working towards more capable and general robotics foundation models.
The long-term vision of this field is to realize generalist robot policies that can readily adapt to new tasks, environments, and embodiments, in a manner more analogous to the open-ended flexibility of human intelligence, in contrast to traditional methods for robotics control.
We believe that our work contributes some modicum of progress towards this goal, which may eventually have profound societal implications in enabling automation to drastically and rapidly replace more human labor.
We feel unqualified to fully assess the potential societal and economic impacts, but think that perhaps, depending on social factors and context, these may range across the spectrum of good to bad, such as aging societies that require more automation to maintain productivity versus the challenges of a labor force made redundant.

\bibliography{main}

@article{intelligence2504pi0,
  title={$\pi$0. 5: a vision-language-action model with open-world generalization, 2025},
  author={Intelligence, Physical and Black, Kevin and Brown, Noah and Darpinian, James and Dhabalia, Karan and Driess, Danny and Esmail, Adnan and Equi, Michael and Finn, Chelsea and Fusai, Niccolo and others},
  journal={URL https://arxiv. org/abs/2504.16054},
  volume={1},
  number={2},
  pages={3},
  year={2025}
}

@article{team2024octo,
  title={Octo: An open-source generalist robot policy},
  author={Team, Octo Model and Ghosh, Dibya and Walke, Homer and Pertsch, Karl and Black, Kevin and Mees, Oier and Dasari, Sudeep and Hejna, Joey and Kreiman, Tobias and Xu, Charles and others},
  journal={arXiv preprint arXiv:2405.12213},
  year={2024}
}

@article{kim2024openvla,
  title={Openvla: An open-source vision-language-action model},
  author={Kim, Moo Jin and Pertsch, Karl and Karamcheti, Siddharth and Xiao, Ted and Balakrishna, Ashwin and Nair, Suraj and Rafailov, Rafael and Foster, Ethan and Lam, Grace and Sanketi, Pannag and others},
  journal={arXiv preprint arXiv:2406.09246},
  year={2024}
}

@inproceedings{o2024open,
  title={Open x-embodiment: Robotic learning datasets and rt-x models: Open x-embodiment collaboration 0},
  author={O’Neill, Abby and Rehman, Abdul and Maddukuri, Abhiram and Gupta, Abhishek and Padalkar, Abhishek and Lee, Abraham and Pooley, Acorn and Gupta, Agrim and Mandlekar, Ajay and Jain, Ajinkya and others},
  booktitle={2024 IEEE International Conference on Robotics and Automation (ICRA)},
  pages={6892--6903},
  year={2024},
  organization={IEEE}
}

@inproceedings{walke2023bridgedata,
  title={Bridgedata v2: A dataset for robot learning at scale},
  author={Walke, Homer Rich and Black, Kevin and Zhao, Tony Z and Vuong, Quan and Zheng, Chongyi and Hansen-Estruch, Philippe and He, Andre Wang and Myers, Vivek and Kim, Moo Jin and Du, Max and others},
  booktitle={Conference on Robot Learning},
  pages={1723--1736},
  year={2023},
  organization={PMLR}
}

@article{khazatsky2024droid,
  title={Droid: A large-scale in-the-wild robot manipulation dataset},
  author={Khazatsky, Alexander and Pertsch, Karl and Nair, Suraj and Balakrishna, Ashwin and Dasari, Sudeep and Karamcheti, Siddharth and Nasiriany, Soroush and Srirama, Mohan Kumar and Chen, Lawrence Yunliang and Ellis, Kirsty and others},
  journal={arXiv preprint arXiv:2403.12945},
  year={2024}
}

@inproceedings{wang2018nervenet,
  title={Nervenet: Learning structured policy with graph neural networks},
  author={Wang, Tingwu and Liao, Renjie and Ba, Jimmy and Fidler, Sanja},
  booktitle={International conference on learning representations},
  year={2018}
}

@inproceedings{huang2020one,
  title={One policy to control them all: Shared modular policies for agent-agnostic control},
  author={Huang, Wenlong and Mordatch, Igor and Pathak, Deepak},
  booktitle={International Conference on Machine Learning},
  pages={4455--4464},
  year={2020},
  organization={PMLR}
}

@article{brohan2022rt,
  title={Rt-1: Robotics transformer for real-world control at scale},
  author={Brohan, Anthony and Brown, Noah and Carbajal, Justice and Chebotar, Yevgen and Dabis, Joseph and Finn, Chelsea and Gopalakrishnan, Keerthana and Hausman, Karol and Herzog, Alex and Hsu, Jasmine and others},
  journal={arXiv preprint arXiv:2212.06817},
  year={2022}
}

@article{doshi2024scaling,
  title={Scaling cross-embodied learning: One policy for manipulation, navigation, locomotion and aviation},
  author={Doshi, Ria and Walke, Homer and Mees, Oier and Dasari, Sudeep and Levine, Sergey},
  journal={arXiv preprint arXiv:2408.11812},
  year={2024}
}

@article{driess2023palm,
  title={Palm-e: An embodied multimodal language model},
  author={Driess, Danny and Xia, Fei and Sajjadi, Mehdi SM and Lynch, Corey and Chowdhery, Aakanksha and Wahid, Ayzaan and Tompson, Jonathan and Vuong, Quan and Yu, Tianhe and Huang, Wenlong and others},
  journal={arXiv preprint arXiv:2303.03378},
  year={2023}
}

@article{ahn2022can,
  title={Do as i can, not as i say: Grounding language in robotic affordances},
  author={Ahn, Michael and Brohan, Anthony and Brown, Noah and Chebotar, Yevgen and Cortes, Omar and David, Byron and Finn, Chelsea and Fu, Chuyuan and Gopalakrishnan, Keerthana and Hausman, Karol and others},
  journal={arXiv preprint arXiv:2204.01691},
  year={2022}
}

@article{ying2021transformers,
  title={Do transformers really perform badly for graph representation?},
  author={Ying, Chengxuan and Cai, Tianle and Luo, Shengjie and Zheng, Shuxin and Ke, Guolin and He, Di and Shen, Yanming and Liu, Tie-Yan},
  journal={Advances in neural information processing systems},
  volume={34},
  pages={28877--28888},
  year={2021}
}

@article{luo2025gcnt,
  title={GCNT: Graph-Based Transformer Policies for Morphology-Agnostic Reinforcement Learning},
  author={Luo, Yingbo and Yao, Meibao and Xiao, Xueming},
  journal={arXiv preprint arXiv:2505.15211},
  year={2025}
}

@article{sferrazza2024body,
  title={Body transformer: Leveraging robot embodiment for policy learning},
  author={Sferrazza, Carmelo and Huang, Dun-Ming and Liu, Fangchen and Lee, Jongmin and Abbeel, Pieter},
  journal={arXiv preprint arXiv:2408.06316},
  year={2024}
}

@inproceedings{perez2018film,
  title={Film: Visual reasoning with a general conditioning layer},
  author={Perez, Ethan and Strub, Florian and De Vries, Harm and Dumoulin, Vincent and Courville, Aaron},
  booktitle={Proceedings of the AAAI conference on artificial intelligence},
  volume={32},
  year={2018}
}

@inproceedings{hong2021structure,
  title={Structure-aware transformer policy for inhomogeneous multi-task reinforcement learning},
  author={Hong, Sunghoon and Yoon, Deunsol and Kim, Kee-Eung},
  booktitle={International Conference on Learning Representations},
  year={2021}
}

@article{buterez2024masked,
  title={Masked attention is all you need for graphs},
  author={Buterez, David and Janet, Jon Paul and Oglic, Dino and Lio, Pietro},
  journal={arXiv preprint arXiv:2402.10793},
  year={2024}
}

@article{velivckovic2017graph,
  title={Graph attention networks},
  author={Veli{\v{c}}kovi{\'c}, Petar and Cucurull, Guillem and Casanova, Arantxa and Romero, Adriana and Lio, Pietro and Bengio, Yoshua},
  journal={arXiv preprint arXiv:1710.10903},
  year={2017}
}

@article{whitman2023learning,
  title={Learning modular robot control policies},
  author={Whitman, Julian and Travers, Matthew and Choset, Howie},
  journal={IEEE Transactions on Robotics},
  volume={39},
  number={5},
  pages={4095--4113},
  year={2023},
  publisher={IEEE}
}

@article{brown2001interval,
  title={Interval estimation for a binomial proportion},
  author={Brown, Lawrence D and Cai, T Tony and DasGupta, Anirban},
  journal={Statistical science},
  volume={16},
  number={2},
  pages={101--133},
  year={2001},
  publisher={Institute of Mathematical Statistics}
}
\bibliographystyle{icml2026}

\appendix
\onecolumn
\raggedbottom

\section{Attention Mask}
\begin{figure}[h]
  \centering
  \includegraphics[width=0.5\columnwidth]{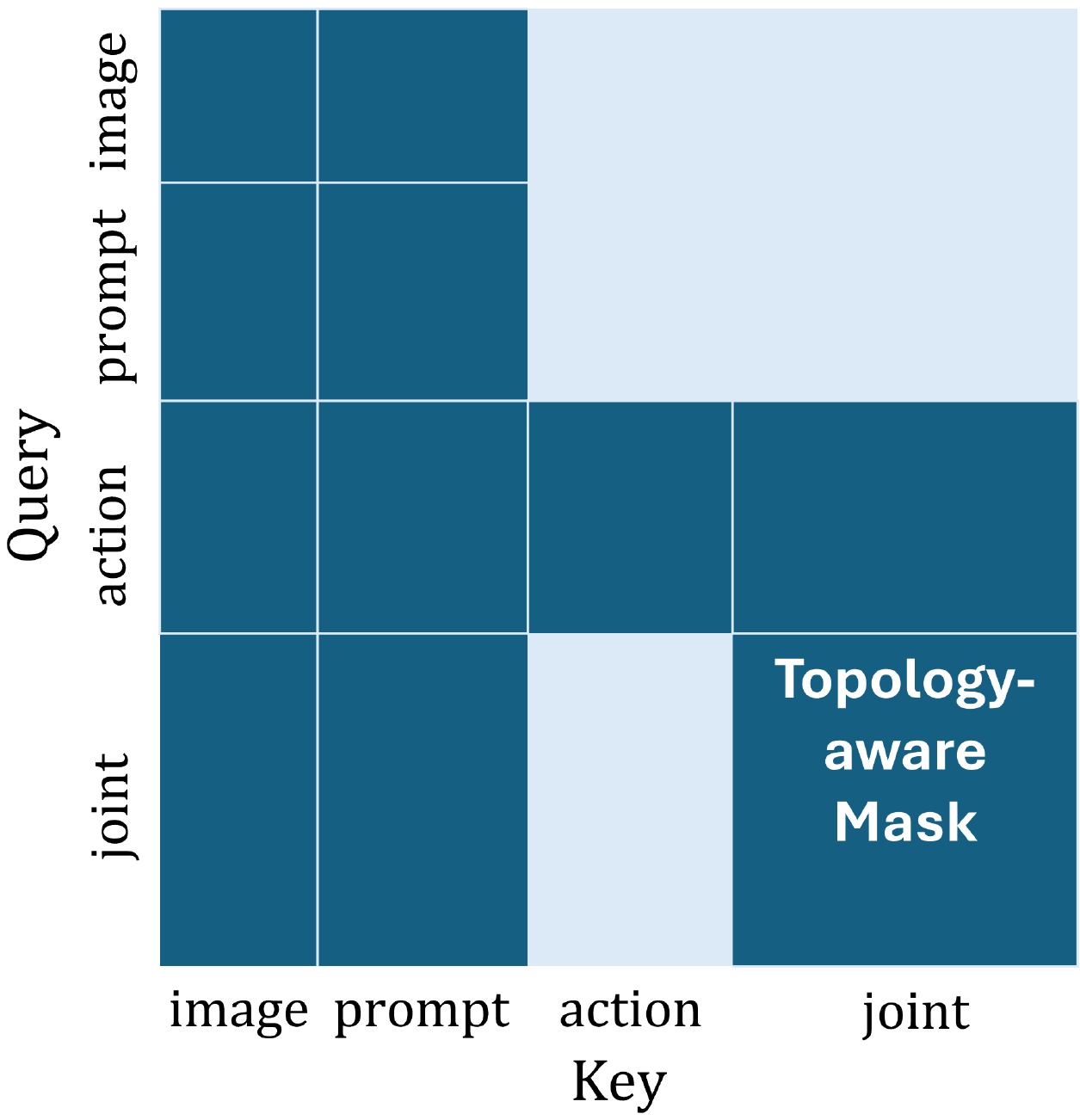}
  \caption{\textbf{Attention mask}:
Attention mask used in our VLA action policy with kinematic tokens.
Dark cells indicate unmasked attention and light cells indicate masked attention.
Tokens are grouped by type for visualization (image/prompt/action/kinematic; each group may contain multiple tokens).
We append kinematic tokens and apply a topology-aware mask only in the joint-to-joint block to encode kinematic connectivity, while all other attention patterns follow the base $\pi_{0.5}$ mask.}

  \label{fig:attention_mask}
\end{figure}

\clearpage

\section{Task Distribution in the DROID 1/8 Subset}
\begin{figure}[h]
  \centering
  \includegraphics[width=1.0\linewidth, trim=0 160 0 200, clip]{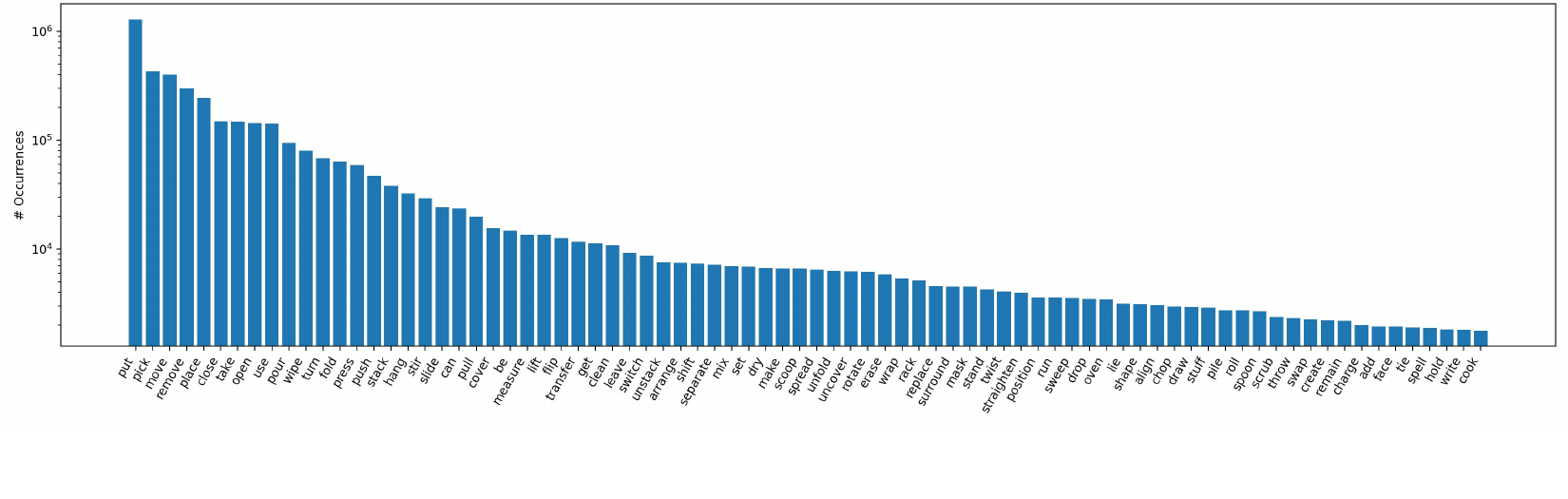}
  \caption{\textbf{Verb distribution in the DROID 1/8 subset}:
Verb frequencies in task instructions.}
\label{fig:subset_verb}
\end{figure}

\begin{figure}[h]
  \centering
  
  \includegraphics[width=1.0\linewidth, trim=0 160 0 180, clip]{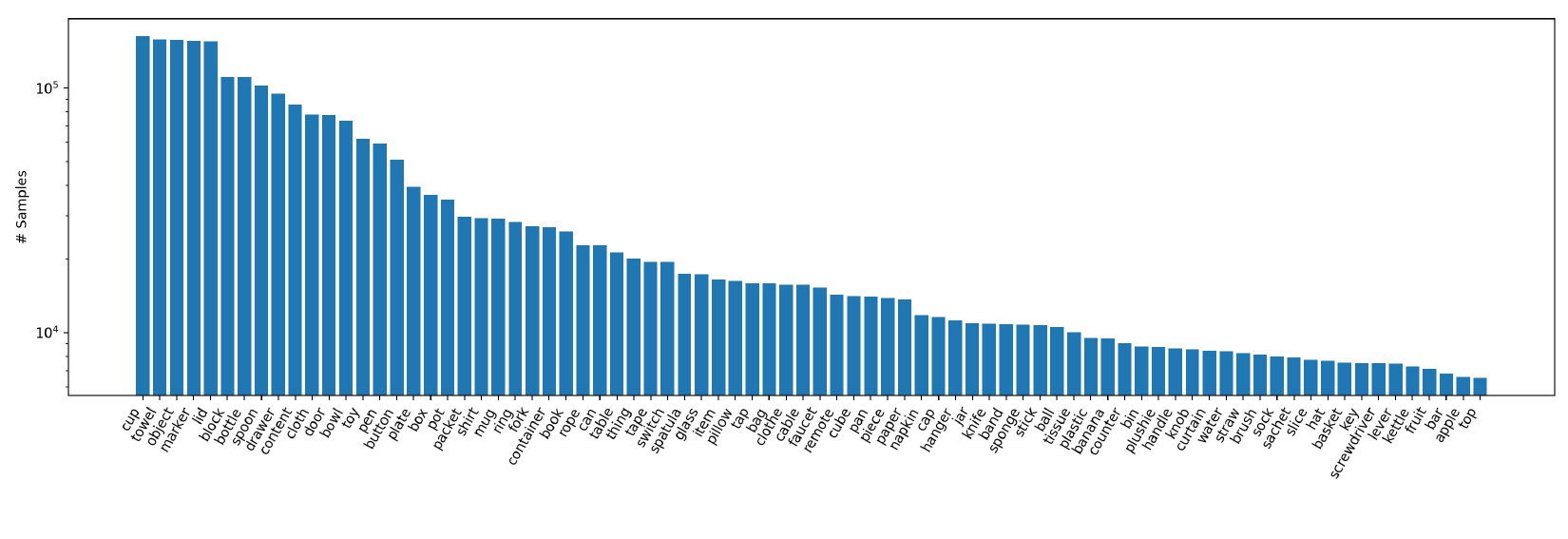}
  \caption{\textbf{Object distribution in the DROID 1/8 subset}:
Object frequencies in task instructions.}
\label{fig:subset_obj}
\end{figure}

We summarize the task distribution of the DROID 1/8 training subset used in our experiments.
We compute frequency statistics of verbs (Figure~\ref{fig:subset_verb}) and objects (Figure~\ref{fig:subset_obj}) appearing in the task instructions.


\FloatBarrier
\clearpage

\section{Training Protocol}
\begingroup
\scriptsize
\setlength{\tabcolsep}{4pt}
\renewcommand{\arraystretch}{1.05}
\setlength\LTleft{0pt}
\setlength\LTright{0pt}

\begin{longtable}{l l >{\raggedright\arraybackslash}p{0.78\linewidth}}
\caption{\textbf{Trainable vs.\ frozen parameters in AP-FT}:
VLM parameters are frozen, while the action-policy parameters are optimized.}
\label{tab:apft_params}\\

\toprule
\textbf{Status} & \textbf{Part} & \textbf{Parameter name} \\
\midrule
\endfirsthead

\toprule
\textbf{Status} & \textbf{Part} & \textbf{Parameter name} \\
\midrule
\endhead

\midrule
\multicolumn{3}{r}{\footnotesize Continued on next page} \\
\endfoot

\bottomrule
\endlastfoot

\multicolumn{3}{l}{\textbf{Frozen: VLM part (2,923,335,408 parameters)}}\\
Frozen & VLM & \paramname{PaliGemma/img/Transformer/encoder_norm/bias} \\
Frozen & VLM & \paramname{PaliGemma/img/Transformer/encoder_norm/scale} \\
Frozen & VLM & \paramname{PaliGemma/img/Transformer/encoderblock/LayerNorm_0/bias} \\
Frozen & VLM & \paramname{PaliGemma/img/Transformer/encoderblock/LayerNorm_0/scale} \\
Frozen & VLM & \paramname{PaliGemma/img/Transformer/encoderblock/LayerNorm_1/bias} \\
Frozen & VLM & \paramname{PaliGemma/img/Transformer/encoderblock/LayerNorm_1/scale} \\
Frozen & VLM & \paramname{PaliGemma/img/Transformer/encoderblock/MlpBlock_0/Dense_0/bias} \\
Frozen & VLM & \paramname{PaliGemma/img/Transformer/encoderblock/MlpBlock_0/Dense_0/kernel} \\
Frozen & VLM & \paramname{PaliGemma/img/Transformer/encoderblock/MlpBlock_0/Dense_1/bias} \\
Frozen & VLM & \paramname{PaliGemma/img/Transformer/encoderblock/MlpBlock_0/Dense_1/kernel} \\
Frozen & VLM & \paramname{PaliGemma/img/Transformer/encoderblock/MultiHeadDotProductAttention_0/key/bias} \\
Frozen & VLM & \paramname{PaliGemma/img/Transformer/encoderblock/MultiHeadDotProductAttention_0/key/kernel} \\
Frozen & VLM & \paramname{PaliGemma/img/Transformer/encoderblock/MultiHeadDotProductAttention_0/out/bias} \\
Frozen & VLM & \paramname{PaliGemma/img/Transformer/encoderblock/MultiHeadDotProductAttention_0/out/kernel} \\
Frozen & VLM & \paramname{PaliGemma/img/Transformer/encoderblock/MultiHeadDotProductAttention_0/query/bias} \\
Frozen & VLM & \paramname{PaliGemma/img/Transformer/encoderblock/MultiHeadDotProductAttention_0/query/kernel} \\
Frozen & VLM & \paramname{PaliGemma/img/Transformer/encoderblock/MultiHeadDotProductAttention_0/value/bias} \\
Frozen & VLM & \paramname{PaliGemma/img/Transformer/encoderblock/MultiHeadDotProductAttention_0/value/kernel} \\
Frozen & VLM & \paramname{PaliGemma/img/embedding/bias} \\
Frozen & VLM & \paramname{PaliGemma/img/embedding/kernel} \\
Frozen & VLM & \paramname{PaliGemma/img/head/bias} \\
Frozen & VLM & \paramname{PaliGemma/img/head/kernel} \\
Frozen & VLM & \paramname{PaliGemma/img/pos_embedding} \\
Frozen & VLM & \paramname{PaliGemma/llm/embedder/input_embedding} \\
Frozen & VLM & \paramname{PaliGemma/llm/final_norm/scale} \\
Frozen & VLM & \paramname{PaliGemma/llm/layers/attn/attn_vec_einsum/w} \\
Frozen & VLM & \paramname{PaliGemma/llm/layers/attn/kv_einsum/w} \\
Frozen & VLM & \paramname{PaliGemma/llm/layers/attn/q_einsum/w} \\
Frozen & VLM & \paramname{PaliGemma/llm/layers/mlp/gating_einsum} \\
Frozen & VLM & \paramname{PaliGemma/llm/layers/mlp/linear} \\
Frozen & VLM & \paramname{PaliGemma/llm/layers/pre_attention_norm/scale} \\
Frozen & VLM & \paramname{PaliGemma/llm/layers/pre_ffw_norm/scale} \\

\midrule

\multicolumn{3}{l}{\textbf{Trainable: Action policy part (430,098,464 parameters)}}\\
Trainable & Action & \paramname{PaliGemma/llm/final_norm_1/Dense_0/bias} \\
Trainable & Action & \paramname{PaliGemma/llm/final_norm_1/Dense_0/kernel} \\
Trainable & Action & \paramname{PaliGemma/llm/layers/attn/attn_vec_einsum_1/w} \\
Trainable & Action & \paramname{PaliGemma/llm/layers/attn/kv_einsum_1/w} \\
Trainable & Action & \paramname{PaliGemma/llm/layers/attn/q_einsum_1/w} \\
Trainable & Action & \paramname{PaliGemma/llm/layers/mlp_1/gating_einsum} \\
Trainable & Action & \paramname{PaliGemma/llm/layers/mlp_1/linear} \\
Trainable & Action & \paramname{PaliGemma/llm/layers/pre_attention_norm_1/Dense_0/bias} \\
Trainable & Action & \paramname{PaliGemma/llm/layers/pre_attention_norm_1/Dense_0/kernel} \\
Trainable & Action & \paramname{PaliGemma/llm/layers/pre_ffw_norm_1/Dense_0/bias} \\
Trainable & Action & \paramname{PaliGemma/llm/layers/pre_ffw_norm_1/Dense_0/kernel} \\
Trainable & Action & \paramname{action_in_proj/bias} \\
Trainable & Action & \paramname{action_in_proj/kernel} \\
Trainable & Action & \paramname{action_out_proj/bias} \\
Trainable & Action & \paramname{action_out_proj/kernel} \\
Trainable & Action & \paramname{time_mlp_in/bias} \\
Trainable & Action & \paramname{time_mlp_in/kernel} \\
Trainable & Action & \paramname{time_mlp_out/bias} \\
Trainable & Action & \paramname{time_mlp_out/kernel} \\

\end{longtable}
\endgroup

\FloatBarrier
\clearpage

\section{Detailed DROID experiments}\label{detailed_DROID}
In addition to success rate (SR), this table reports time-to-success (task completion time) and includes additional ablation variants. Time-to-success is an auxiliary metric and may trade off with SR.
\subsection{Full-/Mix-Mask variants without joint-attribute conditioning (FiLM)}
\begin{table*}[h]
\centering
\scriptsize
\setlength{\tabcolsep}{3pt}
\renewcommand{\arraystretch}{1.08}
\caption{\textbf{DROID simulation evaluation summary}:
Best and second-best are indicated by \textbf{bold} and \uline{underline}, respectively.}
\begin{tabular}{l c c l l r l r r r r r r r r}
\toprule
\textbf{Training} &
\multicolumn{2}{c}{\textbf{Kinematic Token}} &
\textbf{Encoder} &
\textbf{Mask} &
\textbf{Batch} &
\textbf{Train Data} &
\multicolumn{2}{c}{\textbf{Avg}} &
\multicolumn{2}{c}{\textbf{Task 1}} &
\multicolumn{2}{c}{\textbf{Task 2}} &
\multicolumn{2}{c}{\textbf{Task 3}} \\
\cmidrule(lr){2-3}
\cmidrule(lr){8-9}
\cmidrule(lr){10-11}
\cmidrule(lr){12-13}
\cmidrule(lr){14-15}
& \textbf{Chunk} & \textbf{AKT} & & & & &
\textbf{SR(\%)}$\uparrow$ & \textbf{Time[s]}$\downarrow$ &
\textbf{SR(\%)}$\uparrow$ & \textbf{Time[s]}$\downarrow$ &
\textbf{SR(\%)}$\uparrow$ & \textbf{Time[s]}$\downarrow$ &
\textbf{SR(\%)}$\uparrow$ & \textbf{Time[s]}$\downarrow$ \\
\midrule
Full-FT & -- & -- & -- & -- & 256 & Full Dataset & \textbf{66.3} & \uline{3.215} & \textbf{65.7} & \textbf{2.526} & \uline{67.7} & \textbf{4.178} & \textbf{65.7} & \uline{2.939} \\
AP-FT & -- & -- & -- & -- & 32 & 1/8 subset & 19.7 & 4.849 & 18.3 & 4.731 & 15.7 & 6.616 & 25.0 & 3.201 \\
\midrule
AP-FT & 1 & -- & \texttt{linear} & -- & 32 & 1/8 subset & 26.7 & 4.606 & 14.7 & 5.746 & 34.3 & 4.984 & 31.0 & 3.088 \\
AP-FT & 1 & \checkmark & \texttt{linear} & -- & 32 & 1/8 subset & 35.2 & 4.571 & 15.3 & 4.824 & 57.0 & 5.175 & 33.3 & 3.714 \\
AP-FT & 1 & -- & \mbox{\texttt{lin\_swiGlu\_lin}} & -- & 32 & 1/8 subset & 36.0 & 4.925 & 10.3 & 5.450 & \uline{67.7} & 5.254 & 30.0 & 4.072 \\
AP-FT & 1 & \checkmark & \mbox{\texttt{lin\_swiGlu\_lin}} & -- & 32 & 1/8 subset & 37.0 & 4.361 & 11.0 & 4.618 & 60.3 & 5.110 & 39.7 & 3.355 \\
AP-FT & 16 & -- & \texttt{linear} & -- & 32 & 1/8 subset & 16.9 & 5.026 & 14.3 & 5.565 & 16.7 & 6.090 & 19.7 & 3.422 \\
AP-FT & 16 & \checkmark & \texttt{linear} & -- & 32 & 1/8 subset & 18.6 & 5.190 & 8.7 & 5.277 & 27.3 & 5.999 & 19.7 & 4.294 \\
AP-FT & 16 & -- & \mbox{\texttt{lin\_swiGlu\_lin}} & -- & 32 & 1/8 subset & 33.3 & 3.724 & 27.3 & \uline{3.489} & 55.7 & \uline{4.500} & 17.0 & 3.182 \\
AP-FT & 16 & \checkmark & \mbox{\texttt{lin\_swiGlu\_lin}} & -- & 32 & 1/8 subset & 31.0 & 4.876 & 16.7 & 5.664 & 53.7 & 4.748 & 22.7 & 4.216 \\

AP-FT & 1 & -- & \texttt{linear} & Full & 32 & 1/8 subset & 30.5 & 5.317 & 16.0 & 5.378 & 37.0 & 6.541 & 38.0 & 4.033 \\
AP-FT & 1 & \checkmark & \texttt{linear} & Full & 32 & 1/8 subset & 28.0 & 4.824 & 11.0 & 4.556 & 51.0 & 6.018 & 22.0 & 3.899 \\
AP-FT & 1 & -- & \mbox{\texttt{lin\_swiGlu\_lin}} & Full & 32 & 1/8 subset & 30.1 & 4.902 & 15.3 & 4.572 & 34.0 & 6.501 & 41.0 & 3.632 \\
AP-FT & 1 & \checkmark & \mbox{\texttt{lin\_swiGlu\_lin}} & Full & 32 & 1/8 subset & 33.0 & 4.520 & 16.0 & 4.296 & 61.0 & 5.120 & 22.0 & 4.144 \\
AP-FT & 16 & -- & \texttt{linear} & Full & 32 & 1/8 subset & 11.9 & 5.568 & 3.3 & 5.812 & 19.7 & 6.603 & 12.7 & 4.289 \\
AP-FT & 16 & \checkmark & \texttt{linear} & Full & 32 & 1/8 subset & 20.5 & 4.471 & 8.0 & 4.843 & 4.7 & 5.659 & \uline{48.7} & \textbf{2.912} \\
AP-FT & 16 & -- & \mbox{\texttt{lin\_swiGlu\_lin}} & Full & 32 & 1/8 subset & 11.8 & 5.372 & 13.0 & 5.303 & 2.0 & 6.950 & 20.3 & 3.856 \\
AP-FT & 16 & \checkmark & \mbox{\texttt{lin\_swiGlu\_lin}} & Full & 32 & 1/8 subset & 16.9 & 5.203 & 13.7 & 6.026 & 3.0 & 6.258 & 34.0 & 3.326 \\

AP-FT & 1 & -- & \texttt{linear} & Mix & 32 & 1/8 subset & 27.7 & 4.835 & 13.3 & 5.777 & 33.0 & 5.370 & 36.7 & 3.358 \\
AP-FT & 1 & \checkmark & \texttt{linear} & Mix & 32 & 1/8 subset & 35.7 & 4.538 & 22.3 & 4.934 & 44.0 & 5.502 & 40.7 & 3.178 \\
AP-FT & 1 & -- & \mbox{\texttt{lin\_swiGlu\_lin}} & Mix & 32 & 1/8 subset & 36.9 & 4.979 & \uline{27.7} & 4.848 & 56.7 & 5.555 & 26.3 & 4.534 \\
AP-FT & 1 & \checkmark & \mbox{\texttt{lin\_swiGlu\_lin}} & Mix & 32 & 1/8 subset & \uline{47.1} & \textbf{3.117} & 21.0 & 5.010 & \textbf{79.3} & 4.589 & 41.0 & 3.548 \\
AP-FT & 16 & -- & \texttt{linear} & Mix & 32 & 1/8 subset & 21.1 & 5.272 & 0.0 & -- & 21.7 & 6.326 & 41.7 & 4.218 \\
AP-FT & 16 & \checkmark & \texttt{linear} & Mix & 32 & 1/8 subset & 27.2 & 5.105 & 6.3 & 5.807 & 51.7 & 5.916 & 23.7 & 3.591 \\
AP-FT & 16 & -- & \mbox{\texttt{lin\_swiGlu\_lin}} & Mix & 32 & 1/8 subset & 19.0 & 5.284 & 0.0 & 6.240 & 25.0 & 5.949 & 30.3 & 3.662 \\
AP-FT & 16 & \checkmark & \mbox{\texttt{lin\_swiGlu\_lin}} & Mix & 32 & 1/8 subset & 29.6 & \uline{4.849} & 5.3 & 4.510 & 55.3 & 5.689 & 28.0 & 4.348 \\
\bottomrule
\end{tabular}
\end{table*}

\subsection{Full-/Mix-Mask variants with joint-attribute conditioning (FiLM)}
\begin{table*}[h]
\centering
\scriptsize
\setlength{\tabcolsep}{3pt}
\renewcommand{\arraystretch}{1.08}
\caption{\textbf{DROID simulation evaluation summary}:
Best and second-best are indicated by \textbf{bold} and \uline{underline}, respectively.}
\begin{tabular}{l c c l l c r l r r r r r r r r}
\toprule
\textbf{Training} &
\multicolumn{2}{c}{\textbf{Kinematic Token}} &
\textbf{Encoder} &
\textbf{Mask} &
\textbf{FiLM} &
\textbf{Batch} &
\textbf{Train Data} &
\multicolumn{2}{c}{\textbf{Avg}} &
\multicolumn{2}{c}{\textbf{Task 1}} &
\multicolumn{2}{c}{\textbf{Task 2}} &
\multicolumn{2}{c}{\textbf{Task 3}} \\
\cmidrule(lr){2-3}
\cmidrule(lr){9-10}
\cmidrule(lr){11-12}
\cmidrule(lr){13-14}
\cmidrule(lr){15-16}
& \textbf{Chunk} & \textbf{AKT} & & & & & &
\textbf{SR(\%)}$\uparrow$ & \textbf{Time[s]}$\downarrow$ &
\textbf{SR(\%)}$\uparrow$ & \textbf{Time[s]}$\downarrow$ &
\textbf{SR(\%)}$\uparrow$ & \textbf{Time[s]}$\downarrow$ &
\textbf{SR(\%)}$\uparrow$ & \textbf{Time[s]}$\downarrow$ \\
\midrule

AP-FT & -- & -- & -- & -- & -- & 32 & 1/8 subset &
19.7 & 4.849 &
18.3 & 4.731 &
15.7 & 6.616 &
25.0 & 3.201 \\
\midrule

AP-FT & -- & -- & -- & -- & -- & 32 & 1/8 subset &
19.7 & 4.849 &
18.3 & 4.731 &
15.7 & 6.616 &
25.0 & 3.201 \\

AP-FT & 1 & -- & \mbox{\texttt{lin\_swiGlu\_lin}} & -- & -- & 32 & 1/8 subset &
36.7 & 4.747 &
11.3 & 5.802 &
67.3 & 4.597 &
31.3 & 3.843 \\

AP-FT & 1 & -- & \mbox{\texttt{lin\_swiGlu\_lin}} & -- & \checkmark & 32 & 1/8 subset &
37.7 & \uline{3.794} &
6.0 & \textbf{3.727} &
65.0 & \textbf{4.488} &
\uline{42.0} & \uline{3.168} \\

AP-FT & 1 & \checkmark & \mbox{\texttt{lin\_swiGlu\_lin}} & -- & -- & 32 & 1/8 subset &
37.0 & 4.361 &
11.0 & 4.618 &
60.3 & 5.110 &
39.7 & 3.355 \\

AP-FT & 1 & \checkmark & \mbox{\texttt{lin\_swiGlu\_lin}} & -- & \checkmark & 32 & 1/8 subset &
39.4 & 4.245 &
14.6 & \uline{4.502} &
72.7 & 4.606 &
31.0 & 3.626 \\

AP-FT & 1 & -- & \mbox{\texttt{lin\_swiGlu\_lin}} & Mix & -- & 32 & 1/8 subset &
36.3 & 4.881 &
\textbf{27.0} & 4.881 &
54.0 & 5.220 &
28.0 & 4.543 \\

AP-FT & 1 & -- & \mbox{\texttt{lin\_swiGlu\_lin}} & Mix & \checkmark & 32 & 1/8 subset &
\textbf{47.4} & 4.641 &
5.7 & 6.019 &
\uline{77.7} & 4.877 &
\textbf{58.7} & \textbf{3.028} \\

AP-FT & 1 & \checkmark & \mbox{\texttt{lin\_swiGlu\_lin}} & Mix & -- & 32 & 1/8 subset &
\uline{47.1} & \textbf{3.117} &
\uline{21.0} & 5.010 &
\textbf{79.3} & \uline{4.589} &
41.0 & 3.548 \\

\bottomrule
\end{tabular}
\end{table*}

\FloatBarrier
\clearpage

\section{Soft-Mask variants and stability exploration}
\label{app:softmask}

Soft-Mask attention is potentially more expressive than Hard-Mask designs because it retains fully connected attention while injecting topology as an additive bias on the joint-to-joint attention logits.
In practice, however, Soft-Mask can be optimization-sensitive.
Motivated by the bias-initialization ablation in the main text, here we provide a more detailed exploration of Soft-Mask designs, focusing on parameterizations and initialization choices that may improve training stability.
We organize Soft-Mask variants into two families:
(i) \textbf{Adj-SoftMask}, which uses a binary adjacency indicator (connected vs.\ disconnected) and learns a suppression strength; and
(ii) \textbf{SPD-SoftMask}, which uses a shortest-path-distance (SPD) index (Graphormer-style) and learns a distance-indexed bias table (the main-text choice).

\paragraph{Soft-Mask: general form:}
We inject topology into the joint-to-joint attention through an additive bias term
$B^{(\ell)}_{ij}$ applied to the attention logits. Following the main text, we write
\begin{equation}
\alpha^{(\ell)}_{ij}
=
\mathrm{softmax}_{j}\!\left(
\mathrm{logits}^{(\ell)}_{ij}
+
B^{(\ell)}_{ij}
\right).
\label{eq:softmask_general_attn}
\end{equation}
Different Soft-Mask variants correspond to different parameterizations of $B^{(\ell)}_{ij}$.

\subsection{Adj-SoftMask (adjacency + learnable strength)}
Adj-SoftMask parameterizes $B^{(\ell)}_{ij}$ using the binary adjacency indicator $M_{ij}\in\{0,1\}$
defined in the main text ($M_{ij}=1$ for $i=j$ or $(i,j)\in\mathcal{E}$, and $M_{ij}=0$ otherwise).
We instantiate the bias as
\begin{equation}
B^{(\ell)}_{ij}
=
(M_{ij}-1)\, s^{(\ell)}_{ij},
\qquad
s^{(\ell)}_{ij}\in\mathbb{R},
\label{eq:adj_softmask_bias}
\end{equation}
so that $B^{(\ell)}_{ij}=0$ when $M_{ij}=1$ and $B^{(\ell)}_{ij}=-s^{(\ell)}_{ij}$ when $M_{ij}=0$.
Thus, the effect on non-adjacent pairs depends on the sign of $s^{(\ell)}_{ij}$.

\paragraph{Adj-SoftMask (v1.0):}
We instantiate Eq.~\eqref{eq:adj_softmask_bias} with an edge-wise, layer-wise strength
\begin{equation}
s^{(\ell)}_{ij}
=
\exp\!\left(\min\!\left(\theta^{(\ell)}_{ij},\theta_{\max}\right)\right),
\qquad
\theta^{(\ell)}_{ij}\in\mathbb{R}.
\label{eq:adj_softmask_v10}
\end{equation}
Since $s^{(\ell)}_{ij}>0$, the resulting bias on disconnected pairs ($M_{ij}=0$) is always negative.

\paragraph{Adj-SoftMask (v1.1):}
We share the strength across edges within each layer by using a single scalar $\theta^{(\ell)}\in\mathbb{R}$:
\begin{equation}
s^{(\ell)}_{ij}
=
\exp\!\left(\min\!\left(\theta^{(\ell)},\theta_{\max}\right)\right).
\label{eq:adj_softmask_v11}
\end{equation}
Again, $s^{(\ell)}_{ij}>0$ implies a strictly negative bias on disconnected pairs.

\paragraph{Adj-SoftMask (v2.0):}
We use a layer-wise scalar without exponential mapping:
\begin{equation}
s^{(\ell)}_{ij}
=
\theta^{(\ell)},
\qquad
\theta^{(\ell)}\in\mathbb{R}.
\label{eq:adj_softmask_v20}
\end{equation}
Compared to v1.x, this formulation avoids the sharp scaling introduced by $\exp(\cdot)$ and is intended
to improve optimization stability.

\paragraph{Initialization:}
We consider two initializations, matching the main-text bias-initialization taxonomy:
\begin{itemize}
  \item \textbf{Zero}: initialize the effect to be weak (close to \textsc{No-Mask}).
  \item \textbf{Hard}: initialize the effect to be strong (close to \textsc{Hard-Mask}).
\end{itemize}

\subsection{SPD-SoftMask}
Adj-SoftMask uses a binary adjacency signal ($M_{ij}$) and cannot distinguish how far two joints are on the
kinematic graph. SPD-SoftMask (the main-text \textsc{Soft-Mask}) instead indexes $B^{(\ell)}_{ij}$ by the
shortest-path distance (SPD), enabling distance-dependent inductive bias while preserving fully connected attention.

Let $\mathcal{G}=(\mathcal{V},\mathcal{E})$ be the kinematic graph and let $d(i,j)$ denote the SPD between joints
$i$ and $j$ as defined in the main text. 
We parameterize the topology term using a learnable bias table:
\begin{equation}
B^{(\ell)}_{ij} = \theta^{(\ell)}_{d(i,j)},
\qquad
\theta^{(\ell)}_{d}\in\mathbb{R},\ d\in\{0,1,\ldots,D_{\max}\}.
\label{eq:spd_softmask_bias}
\end{equation}
The resulting joint-to-joint attention follows Eq.~\eqref{eq:softmask_general_attn}.

\paragraph{Bias initialization:}
To probe sensitivity to initialization, we consider four initializations of the SPD bias table
$\theta^{(\ell)}_{d}$:
\begin{itemize}
  \item \textbf{Zero}: initialize $\theta^{(\ell)}_{d}=0$ (no topology prior).
  \item \textbf{Hard}: initialize $\theta^{(\ell)}_{d}$ to strongly suppress larger distances (strong locality prior).
  \item \textbf{Mix}: use \textbf{Hard} on even layers and \textbf{Zero} on odd layers.
  \item \textbf{Linear}: initialize $\theta^{(\ell)}_{d}$ with a distance-dependent prior, interpolating from $0$
        to a negative value (e.g., $-3$) as $d$ increases.
\end{itemize}

\paragraph{Warm-start transfer:}
In addition to enabling SPD-SoftMask from \texttt{pi05-base}, we also study a warm-start setting where SPD-SoftMask
is switched on starting from stronger checkpoints:
(i) a \texttt{pi05-base}+JT model (trained with kinematic tokens but without SPD bias), and
(ii) a \texttt{pi05-base}+JT+Mix model (trained with kinematic tokens and Mix-Mask).
We then continue training under the same protocol to evaluate whether SPD-SoftMask can serve as a refinement step
on top of strong masked baselines.

\subsection{Results}
Across these experiments (Table~\ref{tab:softmask_softbot_unified}), we did not find a Soft-Mask configuration that surpasses the Hard-Mask \textbf{Mix-Mask} variant under our training protocol.
Within Adj-SoftMask, v2.0 achieves the strongest results among the tested parameterizations, suggesting that learning the bias magnitude directly can be preferable to exponential mappings.
For SPD-SoftMask, performance is sensitive to the bias initialization and warm-start choice, and we do not observe consistent gains over strong masked baselines.

\renewcommand{\srciPct}[2]{#1{\scriptsize\textcolor{black!75}{\,\,$\pm$\,#2}}}
\begin{table*}[t]
\centering
\caption{\textbf{Soft-Mask results (SR\%)}:
We report per-task success rates (Task1--Task3) in \% along with macro averages, shown as SR\% $\pm$ 95\% CI.}
\scriptsize
\setlength{\tabcolsep}{3pt}
\renewcommand{\arraystretch}{1.12}
\resizebox{\textwidth}{!}{%
\begin{tabular}{l l l r r r r}
\toprule
\textbf{Base Model} & \textbf{Variant} & \textbf{Init.} &
\multicolumn{4}{c}{\textbf{Success Rate (SR\%) (95\% CI)$\uparrow$}} \\
\cmidrule(lr){4-7}
& & & \textbf{Avg} & \textbf{Task1} & \textbf{Task2} & \textbf{Task3} \\
\midrule
pi05-base & Soft-Mask (v1.0) & Zero &
\srciPct{25.9}{4.9} &
\srciPct{26.7}{5.0} &
\srciPct{20.3}{4.5} &
\srciPct{30.7}{5.2} \\
pi05-base & Soft-Mask (v1.0) & Hard &
\srciPct{25.4}{4.9} &
\srciPct{17.7}{4.3} &
\srciPct{15.7}{4.1} &
\srciPct{\uline{42.7}}{5.6} \\
\addlinespace[1pt]

pi05-base & Soft-Mask (v1.1) & Zero &
\srciPct{26.4}{5.0} &
\srciPct{21.3}{4.6} &
\srciPct{16.7}{4.2} &
\srciPct{41.3}{5.5} \\
pi05-base & Soft-Mask (v1.1) & Hard &
\srciPct{\uline{29.5}}{5.1} &
\srciPct{16.7}{4.2} &
\srciPct{\uline{28.7}}{5.1} &
\srciPct{\textbf{43.0}}{5.6} \\
\addlinespace[1pt]

pi05-base & Soft-Mask (v2.0) & Zero &
\srciPct{\textbf{34.4}}{5.3} &
\srciPct{27.6}{5.0} &
\srciPct{\textbf{37.0}}{5.4} &
\srciPct{38.6}{5.5} \\
pi05-base & Soft-Mask (v2.0) & Hard &
\srciPct{23.0}{4.7} &
\srciPct{26.3}{5.0} &
\srciPct{11.6}{3.6} &
\srciPct{31.0}{5.2} \\
\addlinespace[1pt]

pi05-base & Soft-Mask (v3.0) & Zero &
\srciPct{26.1}{4.9} &
\srciPct{\uline{29.7}}{5.1} &
\srciPct{17.3}{4.3} &
\srciPct{31.3}{5.2} \\
pi05-base & Soft-Mask (v3.0) & Hard &
\srciPct{25.1}{4.9} &
\srciPct{18.0}{4.3} &
\srciPct{20.7}{4.6} &
\srciPct{36.7}{5.4} \\
pi05-base & Soft-Mask (v3.0) & Mix &
\srciPct{28.1}{5.1} &
\srciPct{17.0}{4.2} &
\srciPct{24.3}{4.8} &
\srciPct{\textbf{43.0}}{5.6} \\
pi05-base & Soft-Mask (v3.0) & Linear &
\srciPct{20.4}{4.5} &
\srciPct{11.7}{3.6} &
\srciPct{16.7}{4.2} &
\srciPct{32.7}{5.3} \\
\addlinespace[1pt]

pretrained $\pi_{0.5}$ w/ KT & Soft-Mask (v3.0) & -- &
\srciPct{25.8}{4.9} &
\srciPct{23.0}{4.7} &
\srciPct{28.3}{5.1} &
\srciPct{26.0}{4.9} \\
pretrained $\pi_{0.5}$ w/ KT+Mix-Mask & Soft-Mask (v3.0) & -- &
\srciPct{28.9}{5.1} &
\srciPct{\textbf{45.3}}{5.6} &
\srciPct{17.7}{4.3} &
\srciPct{23.7}{4.8} \\
\bottomrule
\end{tabular}%
}
\label{tab:softmask_softbot_unified}
\end{table*}

\FloatBarrier
\clearpage

\section{Multi-embodiment Results}\label{app:multi}
\begin{figure}[h]
  \centering
  \begin{tikzpicture}
    \begin{axis}[
      width=0.7\linewidth,
      height=0.4\linewidth,
      xlabel={Training steps},
      ylabel={Avg SR},
      xmin=0, xmax=125000,
      ymin=0, ymax=0.35,
      xtick={0,50000,80000,125000},
      xticklabels={0,50k,80k,125k},
      scaled x ticks=false,
      ytick={0,0.1,0.2,0.3},
      grid=both,
      major grid style={line width=0.2pt, draw=black!20},
      minor grid style={line width=0.1pt, draw=black!10},
      tick align=outside,
      tick pos=left,
      legend style={at={(0.02,0.98)},anchor=north west,draw=none,fill=none},
      legend cell align=left,
      every axis plot/.append style={line width=1.2pt, mark size=2.6pt},
    ]

      \addplot[name path=bd_upper, draw=none, forget plot]
        coordinates {(0,0.0000) (50000,0.0063) (80000,0.0873) (125000,0.1341)};
      \addplot[name path=bd_lower, draw=none, forget plot]
        coordinates {(0,0.0000) (50000,0.0000) (80000,0.0327) (125000,0.0659)};
      \addplot[orange!70, fill opacity=0.18, draw=none, forget plot]
        fill between[of=bd_upper and bd_lower];

      \addplot[name path=bs_upper, draw=none, forget plot]
        coordinates {(0,0.0000) (50000,0.1596) (80000,0.0185) (125000,0.3338)};
      \addplot[name path=bs_lower, draw=none, forget plot]
        coordinates {(0,0.0000) (50000,0.0404) (80000,0.0000) (125000,0.1662)};
      \addplot[blue!70, fill opacity=0.18, draw=none, forget plot]
        fill between[of=bs_upper and bs_lower];

      \addplot+[color=orange, dashed, mark=o]
        coordinates {(0,0.00) (50000,0.00) (80000,0.06) (125000,0.10)};
      \addlegendentry{$\pi_{0.5}$ (DROID)}

      \addplot+[color=blue, dashed, mark=square]
        coordinates {(0,0.00) (50000,0.10) (80000,0.00) (125000,0.25)};
      \addlegendentry{$\pi_{0.5}$ (SO101)}

      \addplot[name path=od_upper, draw=none, forget plot]
        coordinates {(0,0.0000) (50000,0.2559) (80000,0.2343) (125000,0.2595)};
      \addplot[name path=od_lower, draw=none, forget plot]
        coordinates {(0,0.0000) (50000,0.1641) (80000,0.1457) (125000,0.1671)};
      \addplot[orange!70, fill opacity=0.18, draw=none, forget plot]
        fill between[of=od_upper and od_lower];

      \addplot[name path=os_upper, draw=none, forget plot]
        coordinates {(0,0.0000) (50000,0.1596) (80000,0.2199) (125000,0.2777)};
      \addplot[name path=os_lower, draw=none, forget plot]
        coordinates {(0,0.0000) (50000,0.0404) (80000,0.0801) (125000,0.1223)};
      \addplot[blue!70, fill opacity=0.18, draw=none, forget plot]
        fill between[of=os_upper and os_lower];

      \addplot+[color=orange, solid, mark=triangle]
        coordinates {(0,0.00) (50000,0.21) (80000,0.19) (125000,0.2133)};
      \addlegendentry{Ours (DROID)}

      \addplot+[color=blue, solid, mark=diamond]
        coordinates {(0,0.00) (50000,0.10) (80000,0.15) (125000,0.20)};
      \addlegendentry{Ours (SO101)}

    \end{axis}
  \end{tikzpicture}
  \caption{\textbf{Multi-embodiment learning curves on Panda--SO101}:
  Shaded regions indicate 95\% CI (per-checkpoint).}
  \label{fig:sr_vs_steps}
\end{figure}
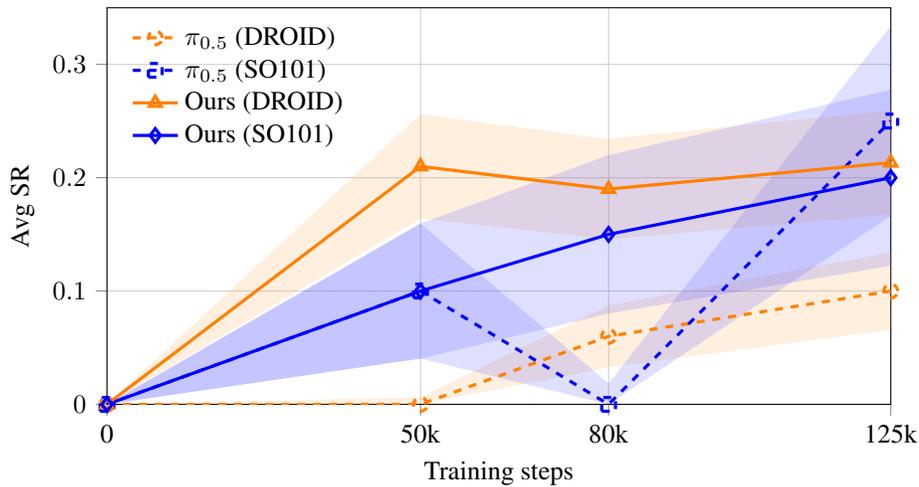

Figure~\ref{fig:sr_vs_steps} reports learning curves for multi-embodiment joint training on the Panda--SO101 mixture, evaluated separately on DROID (Panda) and SO101 in terms of average success rate (Avg SR).
For our method, we use full morphology encoding: kinematic tokens (chunk $G{=}1$), Mix-Mask, and FiLM-based joint-attribute conditioning.
Overall, our embodiment-aware policy improves performance on DROID throughout training, while achieving comparable performance on SO101.

\textbf{DROID (Panda):}
Our method improves earlier and remains higher: at 50k steps it reaches Avg SR 0.210 ($\pi_{0.5}$: 0.000), and at 125k steps it remains higher at 0.213 ($\pi_{0.5}$: 0.100).

\textbf{SO101:}
Performance is comparable across training: at 50k steps both reach Avg SR 0.100, and at 125k steps our method achieves 0.200 while $\pi_{0.5}$ achieves 0.250 (difference: 0.050).

We hypothesize that the larger gains on DROID are due to the training setup and the mixture ratio being skewed toward Panda, which can favor improvements on DROID relative to SO101. Under AP-FT, single-embodiment SO101 training with $\pi_{0.5}$ reaches only about Avg SR 0.2, suggesting that SO101 performance is generally more constrained in this fine-tuning regime.

\clearpage

\end{document}